\documentclass{article}

\usepackage{microtype}
\usepackage{graphicx}
\usepackage{subfigure}
\usepackage{booktabs}
\usepackage[T1]{fontenc}
\usepackage{siunitx}
\usepackage{ulem}
\usepackage[utf8]{inputenc}
\usepackage{mdframed}  
\usepackage{dingbat}
\usepackage{xcolor}
\usepackage{wrapfig}
\usepackage{hyperref}
\usepackage{multirow}
\usepackage{floatflt}

\usepackage[accepted]{icml2025}

\usepackage{amsmath}
\usepackage{amssymb}
\usepackage{mathtools}
\usepackage{amsthm}
\usepackage{longtable}

\usepackage[capitalize,noabbrev]{cleveref}

\theoremstyle{plain}

\theoremstyle{definition}

\theoremstyle{remark}

\usepackage[textsize=tiny]{todonotes}

\icmltitlerunning{Joint Scaling Laws for Mixture of Experts}

\newcommand{\dm}{d_{\text{model}}}

\newcommand{\dv}{d_{\text{vocab}}}
\newcommand{\Nact}{N_{\text{act}}}
\newcommand{\Ntotal}{N_{\text{total}}}
\newcommand{\hE}{\hat{E}}
\newcommand{\Nane}{N_{act\setminus e}}

\DeclareMathOperator*{\argmin}{arg\,min}

\renewcommand{\log}{\ln}

\newcommand{\round}[1]{\num[round-mode=places,round-precision=4]{#1}}

\definecolor{purp}{rgb}{0.58, 0.0, 0.83}

\definecolor{e1}{rgb}{0.0,0.0,0.0}
\definecolor{e1}{rgb}{0.0,0.0,0.0}
\definecolor{e2}{rgb}{0.0,0.0,0.0}
\definecolor{e4}{rgb}{0.0,0.0,0.0}
\definecolor{e8}{rgb}{0.0,0.0,0.0}
\definecolor{e16}{rgb}{0.0,0.0,0.0}
\definecolor{e32}{rgb}{0.0,0.0,0.0}

\begin{document}

\onecolumn

\icmltitle{Joint MoE Scaling Laws:\\Mixture of Experts Can Be Memory Efficient}

\icmlsetsymbol{core}{*}

\begin{icmlauthorlist}
\icmlauthor{Jan Ludziejewski}{core,uw,ideas}
\icmlauthor{Maciej Pióro}{core,ideas,ipptpan}
\icmlauthor{Jakub Krajewski}{core,uw,ideas} \\
\icmlauthor{Maciej Stefaniak}{uw}
\icmlauthor{Michał Krutul}{uw,ideas}
\icmlauthor{Jan Małaśnicki}{uw,ideas} \\
\icmlauthor{Marek Cygan}{uw,nomagic}
\icmlauthor{Piotr Sankowski}{uw,mimsolutions}
\icmlauthor{Kamil Adamczewski}{ideas,wroc} \\
\icmlauthor{Piotr Miłoś}{ideas,impan}
\icmlauthor{Sebastian Jaszczur}{uw,ideas}
\end{icmlauthorlist}

\icmlaffiliation{ideas}{IDEAS NCBR}
\icmlaffiliation{mimsolutions}{MIM Solutions}
\icmlaffiliation{uw}{University of Warsaw}
\icmlaffiliation{nomagic}{Nomagic}
\icmlaffiliation{wroc}{Wroclaw University of Science and Technology}
\icmlaffiliation{impan}{Institute of Mathematics, Polish Academy of Sciences}
\icmlaffiliation{ipptpan}{Institute of Fundamental Technological Research, Polish Academy of Sciences}

\icmlcorrespondingauthor{Jan Ludziejewski}{ludziej@mimuw.edu.pl}
\icmlcorrespondingauthor{Maciej Pióro}{maciej.pioro@gmail.com}
\icmlcorrespondingauthor{Jakub Krajewski}{gim.jakubk@gmail.com}
\icmlcorrespondingauthor{Sebastian Jaszczur}{sebastian.jaszczur@gmail.com}

\icmlkeywords{Machine Learning, ICML}

\vskip 0.3in

\printAffiliationsAndNotice{\icmlCoreContribution}

\begin{abstract}

\looseness=-1 Mixture of Experts (MoE) architectures have significantly increased computational efficiency in both research and real-world applications of large-scale machine learning models. However, their scalability and efficiency under memory constraints remain relatively underexplored. In this work, we present joint scaling laws for dense and MoE models, incorporating key factors such as the number of active parameters, dataset size, and the number of experts. Our findings provide a principled framework for selecting the optimal MoE configuration under fixed memory and compute budgets. Surprisingly, we show that MoE models can be more memory-efficient than dense models, contradicting conventional wisdom. To derive and validate the theoretical predictions of our scaling laws, we conduct over $280$ experiments with up to $2.7$B active parameters and up to $5$B total parameters. These results offer actionable insights for designing and deploying MoE models in practical large-scale training scenarios.

\end{abstract}

\vspace{0.2cm}
\section{Introduction}

Recently, language models have grown increasingly large, a trend accelerated by Mixture of Experts (MoE) techniques~\citep{fedus2022switch,du2022glam}. MoE models are now widely adopted~\citep{jiang2024mixtral, dai2024deepseekmoe} and are generally considered compute-efficient~\citep{ludziejewskiscaling, clark2022unified}, though often considered memory-inefficient~\cite{zadouri2023pushing}. However, the precise trade-offs between compute and memory efficiency have remained unclear so far. 

Consider a motivating question: Can an MoE model be the optimal choice when constrained by a fixed memory budget, such as a single H100 node? Increasing the number of experts has a relatively minimal impact on the cost in FLOPs but can drastically increase memory requirements, often to prohibitive levels depending on the specific hardware and load.

In order to answer this question, we derive a \textit{joint} scaling law for both dense and MoE models, accounting for key factors such as the number of active parameters, dataset size, and number of experts. This framework provides a rigorous analysis of model performance under strict memory constraints. Our findings reveal that, contrary to common assumptions, MoE models can be more memory-efficient than dense models---that is, MoE models with the same loss and training budget can have lower memory usage than dense models.

Our work is the first to provide detailed guidance on selecting the optimal number of experts for MoE models, balancing computational budget and memory. Our conclusions are based on extensive large-scale experiments with over $280$ models, scaled up to $2.7$B active parameters and up to $5$B total parameters\footnote[2]{We plan to open-source model checkpoints and the code.}. For a complete list of experiments, see Appendix~\ref{sec:full_experiments}.

\newpage

\begin{figure}[H]
    \centering
    \subfigure[]{
        \includegraphics[width=0.48\textwidth]{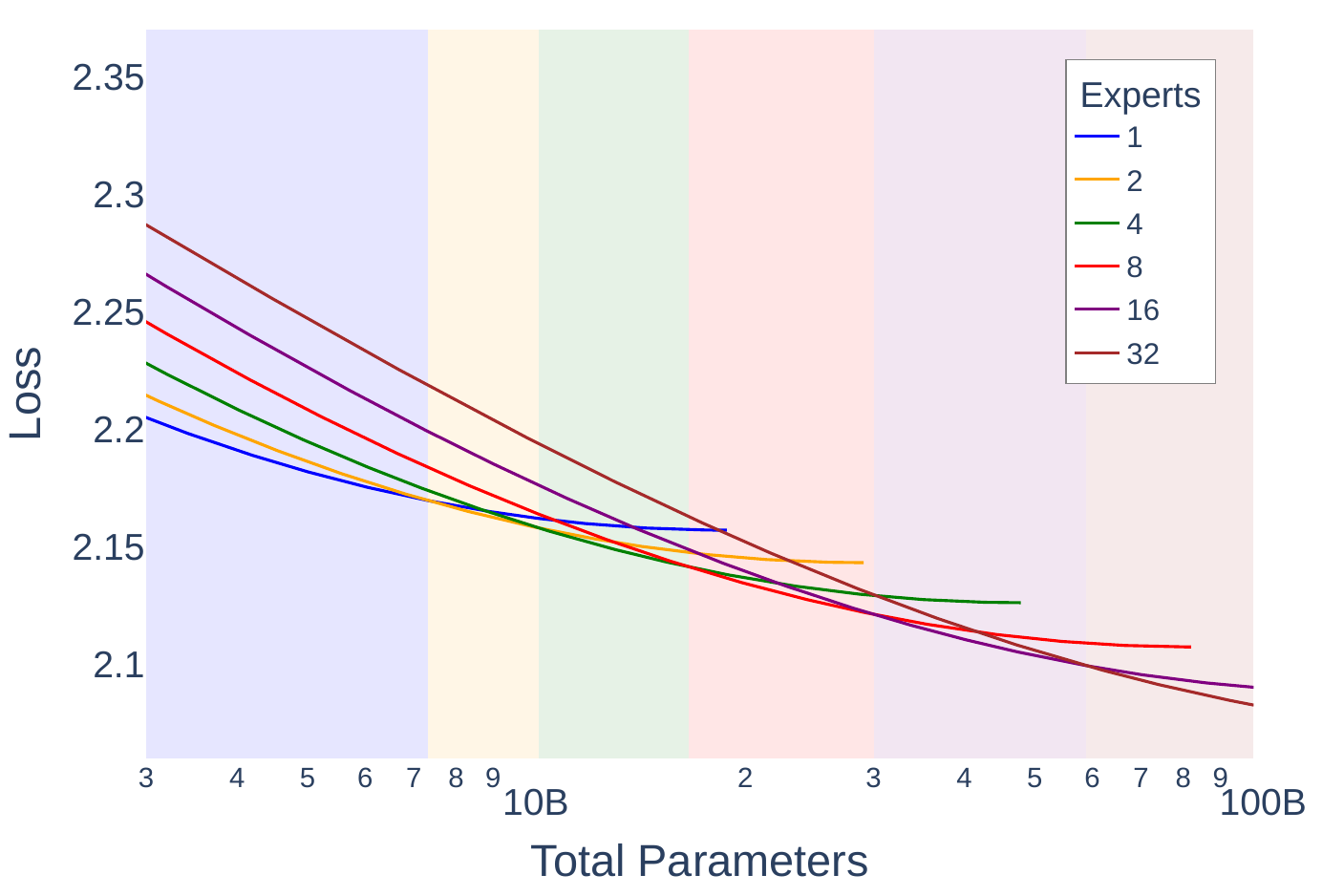}
    }
    \subfigure[]{
        \includegraphics[width=0.48\textwidth]{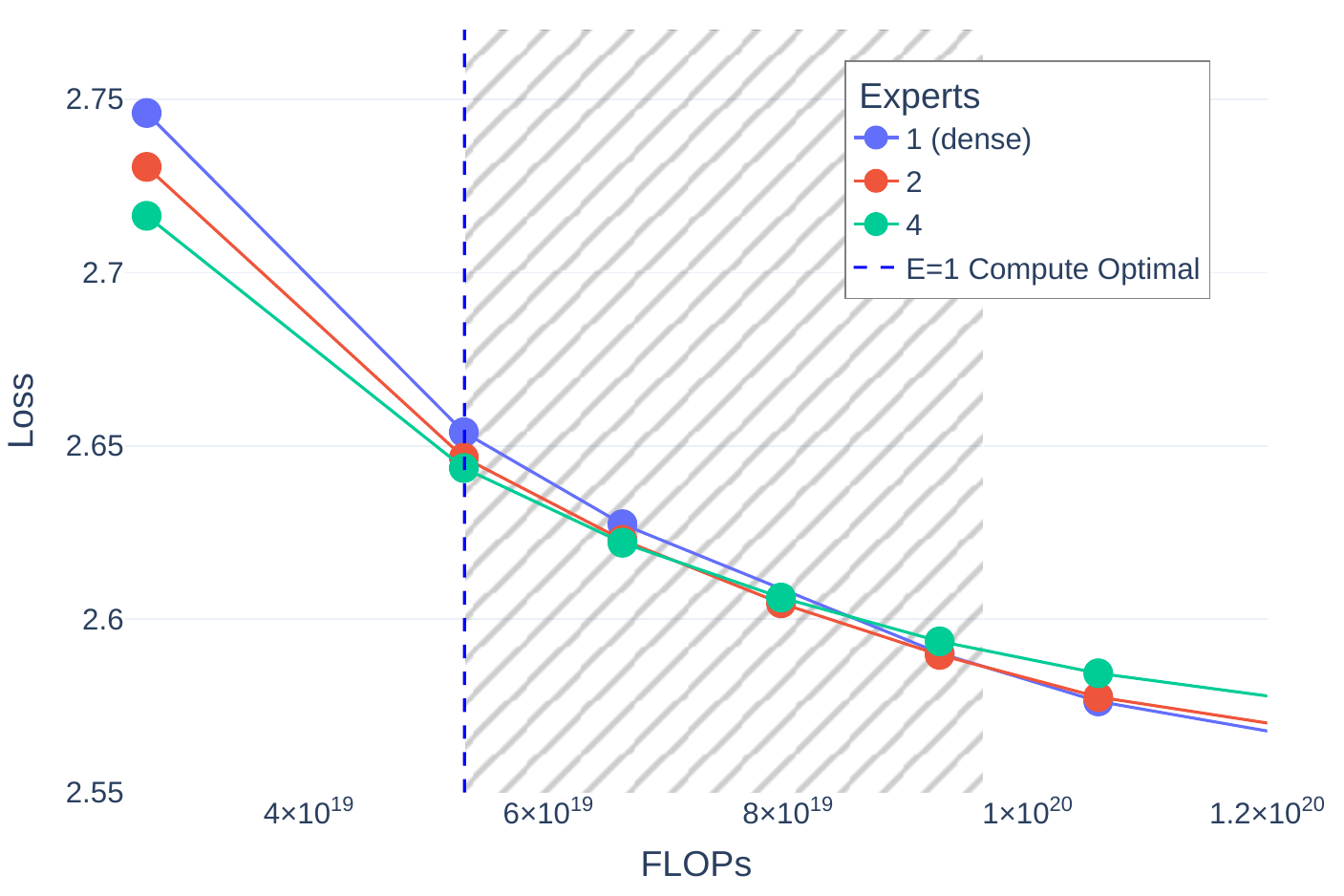}
    }
\caption{\textbf{(a)} The loss of memory-constrained models predicted using our scaling law under a fixed training budget of $10^{22}$ FLOPs. Each curve represents a different number of experts. The lines are truncated at compute-optimal points since undertrained models are both larger and worse in terms of loss, thus pointless in a memory-constrained scenario. Shaded areas indicate the memory-optimal number of experts for the corresponding memory budgets.
\textbf{(b)} Experimental validation of the thesis that MoE can be memory-optimal. The marked area shows an interval in which training a compute-matched MoE achieves better loss than an overtrained dense model with the same number of total parameters ($1.1$B). Such an MoE is trained for longer and has fewer active parameters, making it more practical for inference.}   
  \label{fig:thesis_proof}
\end{figure}

In summary, the key contributions of this work are:
\begin{itemize}
    \item We derive a joint scaling law for Mixture of Experts and dense models,
    \vspace{-0.1cm}
    \begin{align}
    \mathcal{L}(\Nact, D, \hE) = \;a{\hE}^{\delta}\Nact^{\alpha + {\gamma}{\log}(\hE)} 
     +  \;b{\hE}^{\omega}D^{\beta + {\zeta}{\log}(\hE)} + c,
    \end{align}

    \vspace{-0.35cm}
    where $\mathcal{L}$ represents the final training loss, $\Nact$ denotes the number of active parameters, $D$ is the dataset size, $\hE$ is a monotonic transformation of the number of experts (as defined in Equation~\eqref{eq:e_transformation}), and $c$ is the minimum achievable loss on the dataset, often called the irreducible entropy of the dataset.
    \item Based on the proposed scaling law, we show that the choice of the optimal number of experts (including dense models with $E=1$) depends on specific computational and memory constraints, see Figure~\ref{fig:thesis_proof}. Furthermore, we demonstrate how the optimal token-to-parameter ratio depends on $E$. 
    \item We show that MoE can often be the preferred alternative to dense models, even if GPU memory is the constraining factor. We validate our theoretical findings by training a set of $1.1$B-parameter models under identical compute and total memory budgets. The MoE models achieve a lower final loss, confirming their superior efficiency in practice. Moreover, we observe that MoE models not only have lower loss but also deliver higher performance during inference.
\end{itemize}

\section{Related Work}

\textbf{Mixture of Experts.}
Mixture of Experts (MoE) was introduced by \citet{moe1991}, who combined a gating network with a set of expert networks. \citet{shazeer2017outrageously} applied MoE to an LSTM-based model~\citep{hochreiter1997long}, scaling the architecture up to $137$ billion parameters. In Transformer-based LLMs, MoE is most often applied as a replacement for the feed-forward layer~\citep{lepikhin2020gshard, shazeer2018meshtensorflow}. It replaces the feed-forward layer's MLP with a set of expert MLPs along with a router, which selects one or more MLPs for each token. With the recent surge in LLM research, MoE models are gaining even more traction. This is exemplified by the development of extremely large-scale models such as DeepSeek-R1 and Qwen2.5-Max~\citep{deepseekr1, qwen25}.
In our work, we use the standard Switch MoE layer~\citep{fedus2022switch}, which routes each token to one expert and encourages even token-to-expert assignment via the addition of a differentiable load-balancing loss.

\newpage
\textbf{Scaling Laws.}  
Scaling laws refer to empirically derived equations that relate model loss
to factors such as the number of parameters, the quantity of training data, or the computational budget. For dense Transformers, scaling laws were initially explored by \citet{hestness2017deep} and \citet{kaplan2020scaling}, who identified power-law relationships between the final loss, model size, and dataset size. \citet{hoffmann2022training} expanded these by incorporating variable cosine cycle lengths and adjusting the functional form of the equation:
\vspace{-0.1cm}
\begin{equation} \label{eq:scaling_law_chinchilla}
  \mathcal{L}(\Nact, D) = m\Nact^{\mu} + nD^{\nu} + c.
\end{equation}

Scaling laws have also been applied to other architectures and training setups. \citet{henighan2020scaling} examined autoregressive modeling across multiple modalities, while \citet{ghorbani2021scaling} focused on machine translation. \citet{frantar2023scaling} studied the effects of pruning on vision and language Transformers, determining optimal sparsity given a fixed compute budget.

\begin{figure*}
    \centering
    \subfigure[]{
        \includegraphics[width=0.48\textwidth]{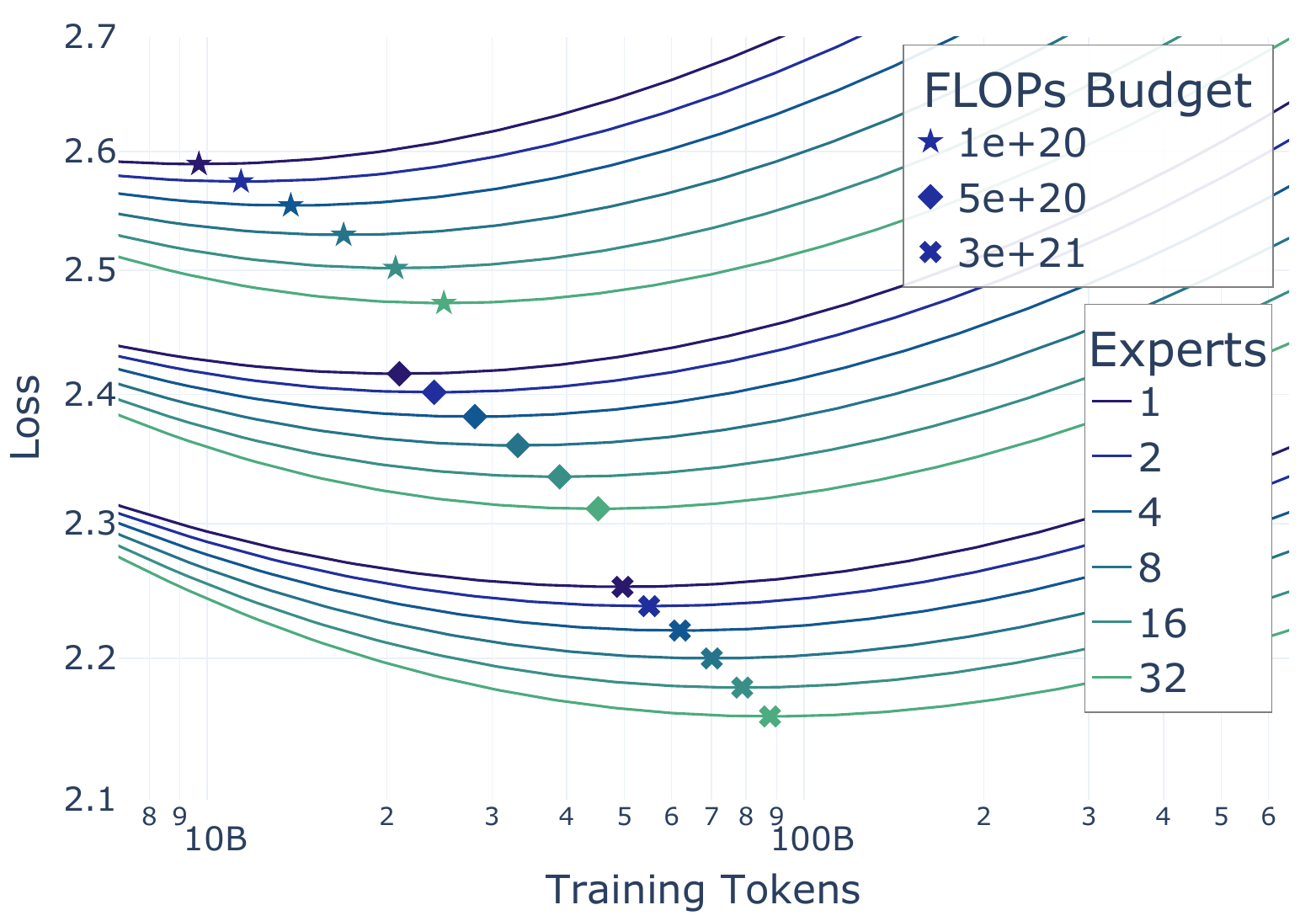}
    }
    \subfigure[]{
        \includegraphics[width=0.48\textwidth]{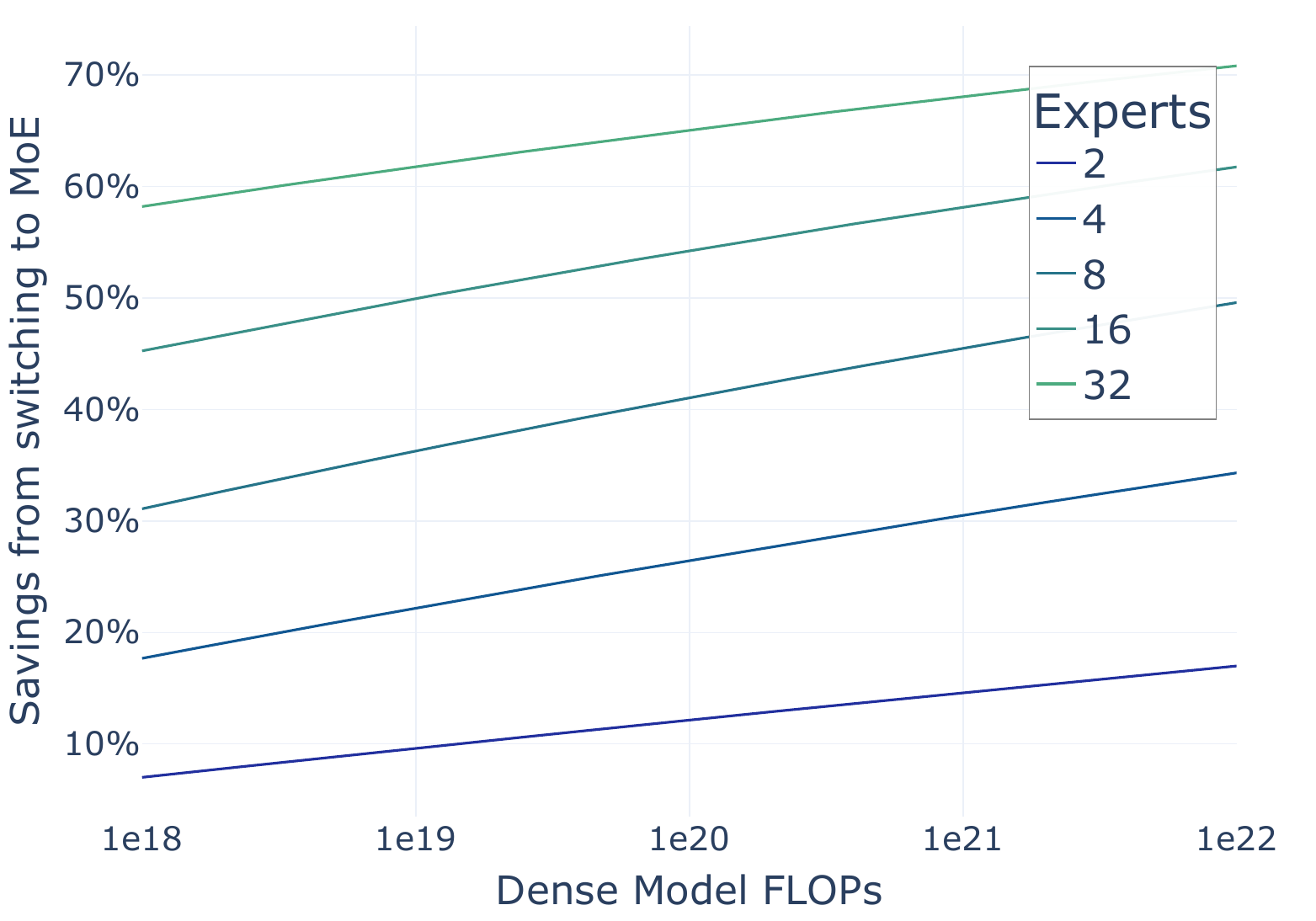}
    }
\caption{\textbf{(a)} IsoFLOP profiles for selected training budgets, with compute-optimal points marked for each curve. \textbf{(b)} FLOP savings from switching from a compute-optimal dense model to a compute-optimal MoE. For instance, 40\% savings at $1$e$20$ FLOPs mean that an MoE matching the performance of a compute-optimal dense model trained with $1$e$20$ FLOPs can be trained with just $6$e$19$ FLOPs (60\% of the dense's budget). The advantage of using MoE increases with larger models and expert counts.}
    
  \label{fig:isoflops}
\end{figure*}

\citet{clark2022unified} investigated scaling in MoE models, varying model size and the number of experts on a fixed dataset. They concluded that routed models are more efficient only up to a certain size. Their formula took the form:
\begin{equation} \label{eq:scaling_law_clark}
  \mathcal{L}(\Nact, \hE) = a\hE^{\delta}\Nact^{\alpha + \gamma \log(\hE)},
\end{equation}
where $\hE$ is a monotonic transformation of the number of experts $E$, defined as:
\begin{equation} \label{eq:e_transformation}
    \frac{1}{\hE} = \frac{1}{E - 1 + \left(\frac{1}{E_{\text{start}}} - \frac{1}{E_{\text{max}}}\right)^{-1}} + \frac{1}{E_{\text{max}}}.
\end{equation}
These analyses have since been extended by \citet{ludziejewskiscaling} and \citet{dai2024deepseekmoe}, who considered variable dataset size as well as the granularity of experts. In our work, we keep the experts non-granular; however, we treat the number of experts and the number of training tokens as variables. \citet{sardana2024chinchillaoptimalaccountinginferencelanguage} assumes a fixed joint inference and training budget. We make similar assumptions; however, we consider accelerator memory as a limiting factor and extend the analysis to MoE models, which can serve as a more compute-friendly alternative to dense models. \citet{yun2024inferenceoptimalmixtureofexpertlargelanguage} have focused on MoE inference optimality and measuring real hardware efficiency.

Concurrently to our work, \citet{abnar2025parametersvsflopsscaling} derived scaling laws for optimal sparsity while considering the interplay between training FLOPs and model size. They also investigated the relationship between pretraining loss and downstream performance, noting differences between MoE and dense models on certain tasks. In contrast, we analyze not only training FLOPs and model size but also inference cost and total memory usage. Additionally, we derive and utilize a principled method for scaling the learning rate with the number of experts and model size, along with describing further adjustments to enable researchers to use scaling laws economically and reliably.

\newpage
\section{Joint MoE Scaling Laws}

We now derive the functional form of our joint scaling laws for both dense Transformers and MoE, relating the number of active model parameters $\Nact$, training tokens $D$, and MoE experts $E$. 

\textbf{Fixed Number of Experts.} Following \citet{hoffmann2022training} and established practices in the literature~\citep{frantar2023scaling, kumar2024scalinglawsprecision, ludziejewskiscaling}, we postulate the following form of the equation:
\begin{align}\label{eq:general_form}
\mathcal{L}(\Nact, D, E) = m(E) \Nact^{\mu(E)} \hspace{-0.07cm} + n(E) D^{\nu(E)} \hspace{-0.02cm} + c(E),
\end{align}
assuming that if we fix the number of experts, the model's performance can be described using Equation~\ref{eq:scaling_law_chinchilla}. In the subsequent part, we will postulate how $m, \mu, n, \nu, c$ depend on $E$, deriving the joint equation.

\textbf{Constant Factor.} $c(E)$ represents irreducible loss caused by the inherent entropy of the dataset. Thus, it does not depend on the architecture ($E$ in our case):
\begin{align*}
    c(E) := c.
\end{align*}
\vspace{-0.3cm}

\textbf{Interaction of $\boldsymbol{E}$ with Model and Dataset Size.} 
To quantify the interaction between the number of experts and other training parameters, we gather observations from related work:
\vspace{-0.3cm}
\begin{enumerate}
    \item Scaling the number of experts ($E$) can be described as a power law~\citep{clark2022unified}.
    \item For a fixed number of training tokens ($D$), as model size ($\Nact$) increases, the benefit of using an MoE diminishes~\citep{clark2022unified}.
    \item For a fixed model size ($\Nact$), as the number of training tokens increases, the benefit of an MoE grows~\citep{ludziejewskiscaling}.
\end{enumerate}
Motivated by Observation 1, we set 
\vspace{-0.1cm}
\begin{align*}
    m(E)=aE^\delta, \quad n(E)=bE^\omega,
\end{align*}
reflecting the power-law relation between $E$ and the loss.

Additionally, to ensure flexibility in modeling Observations 2 and 3, we introduce an interaction with the exponents over $\Nact$ and $D$:
\vspace{-0.3cm}
\begin{align*}
    \mu(E) &=\alpha+\gamma\log(E), \\ 
    \nu(E) &= \beta+\zeta\log(E).
\end{align*}
Note that if we ignore the second and third terms in Equation~\ref{eq:general_form}, this yields a functional form identical to Equation~\ref{eq:scaling_law_clark}. 

Empirically, we observe a good fit for our formula, as described in Section~\ref{sec: fitting_scaling_law}. This shows that our proposed interactions between $E$, $\Nact$, and $D$ can accurately model the performance of MoE models.

\looseness=-1\textbf{Modeling of $\boldsymbol{E}$.} When the number of experts is small, a certain overhead—caused, for example, by interference from auxiliary losses—can overshadow the benefits of conditional computation. Additionally, employing a very large number of experts brings diminishing returns. To address these phenomena, we follow \citet{clark2022unified} and use a transformation of the number of experts $\hE$ as given in Equation~\ref{eq:e_transformation}.

\textbf{Joint MoE Scaling Law.} By combining these observations, we establish the final form of our scaling law:
\begin{equation} \label{eq:scaling_law_final} 
  \mathcal{L}(\Nact, D, \hE) = \\
  {a{\hE}^{\delta}}\Nact^{\alpha + {\gamma}{\log}(\hE)} + {b{\hE}^{\omega}}D^{\beta + {\zeta}{\log}(\hE)} + c. 
\end{equation}

We fit the coefficients in Equation~\ref{eq:scaling_law_final} based on the results of our experiments; see Table~\ref{tab:scaling_law_params}. In Section~\ref{sec:optimal_models}, we present the outcomes and findings derived from the scaling laws. The details of the training runs, as well as the fitting procedure, are described in Section~\ref{sec: fitting_scaling_law}.

\newpage
\section{Compute and Memory Optimality}
\label{sec:optimal_models}

\looseness=-1 In this section, we employ our scaling laws to offer recommendations for optimal configurations in different training and inference scenarios. Refer to Appendix~\ref{app:technical} for details on counting FLOPs, the relationships between active and total parameters, and other technical aspects.

\subsection{Compute Optimality}

\begin{wrapfigure}{r}{0.55\textwidth}
\vspace{-0.5cm}
\begin{minipage}{\linewidth}
\begin{mdframed}[
  backgroundcolor=gray!20,
  linecolor=black,
  linewidth=0pt,
  roundcorner=5pt,
  innertopmargin=1em,
  innerbottommargin=1em,
  nobreak=true
]
\textbf{Finding 1.}
\textbf{More experts $\boldsymbol{\rightarrow}$ higher tokens-to-param ratio.} \\ 
Assume a fixed compute budget. In this scenario, when increasing the number of experts, it is optimal to decrease the number of active parameters and increase the number of training tokens accordingly (Table~\ref{tab:opt_n_d_table}).

\end{mdframed}
\end{minipage}
\vspace{-1cm}
\end{wrapfigure}

A model is considered compute-optimal if it achieves the lowest loss among models trained with the same compute budget $F$. To find such an optimal configuration, we optimize the following:
\begin{align*}
  &\argmin_{\Nact, D, E} \mathcal{L}(\Nact, D, E) \\
    &\text{s.t. } 6 \Nact D = F
\end{align*}

\textbf{Optimal $\boldsymbol{N}$ and $\boldsymbol{D}$ Depend on the Number of Experts.} 
Assuming a given number of experts $E$, the compute-optimal training configuration can be achieved by selecting the appropriate trade-off between training tokens and model size. IsoFLOP slices comparing the predicted loss with dataset size for selected compute budgets are plotted in Figure~\ref{fig:isoflops}~(a). 

For any fixed $E$, our scaling law has the Chinchilla functional form of Equation~\ref{eq:scaling_law_chinchilla}. Thus, from \citet{hoffmann2022training}, the compute-optimal number of tokens and active parameters for the budget $F$ and the number of experts $E$ are given by
\vspace{-0.15cm}
\begin{align}
    N^{\text{opt}}_{\text{act}}(F) = G \left( \frac{F}{6} \right)^a, \quad D^{\text{opt}}(F) = G^{-1}\left(\frac{F}{6}\right)^b,
\end{align}
\vspace{-0.2cm}
where
\vspace{-0.15cm}
\begin{align*}
    G = \left(\frac{\mu(E)m(E)}{\nu(E)n(E)}\right)^{\frac{1}{\mu(E)+\nu(E)}}, \quad
    a = \frac{\nu(E)}{\mu(E)+\nu(E)}, \quad b = \frac{\mu(E)}{\mu(E)+\nu(E)}.
\end{align*}

\begin{wraptable}{r}{0.55\textwidth}
\vspace{-0.35cm}
\centering
\label{tab:example}
\begin{tabular}{lccc}
\toprule
\textbf{Training Budget} & \textbf{Experts} & \textbf{$N_\text{act}^\text{opt}$} & \textbf{$D^\text{opt}$}\\
\midrule
\multirow[t]{5}{*}{$10^{20}$} 
 & $1$   & $1.7$B  & $9.7$B   \\
 & $2$   & $1.5$B  & $11.4$B  \\
 & $4$   & $1.2$B  & $13.9$B  \\
 & $8$   & $990$M  & $17$B    \\
 & $16$  & $810$M  & $20.7$B  \\
 & $32$  & $669$M  & $24.9$B  \\
\cmidrule(lr){1-4}
\multirow[t]{5}{*}{$10^{21}$} 
 & $1$   & $5.7$B  & $29.3$B  \\
 & $2$   & $5$B & $33$B \\
 & $4$   & $4.4$B & $38$B \\
 & $8$   & $3.8$B & $44.3$B       \\
 & $16$  & $3.3$B  & $51.2$B   \\
 & $32$  & $2.85$B  & $58.4$B  \\
\cmidrule(lr){1-4}
\multirow[t]{5}{*}{$10^{22}$} 
 & $1$   & $18.8$B  & $88.6$B   \\
 & $2$   & $17.4$B  & $96$B  \\
 & $4$   & $15.8$B  & $105.4$B  \\
 & $8$   & $14.4$B  & $115.8$B    \\
 & $16$  & $13.2$B  & $126.5$B  \\
 & $32$  & $12.2$B  & $136.9$B  \\
\bottomrule
\end{tabular}
\caption{Example compute-optimal training configurations for MoE models. For every training budget, as the number of experts increases, the optimal $D^\text{opt}$ also increases while $N_\text{act}^\text{opt}$ decreases.}
\label{tab:opt_n_d_table}
\vspace{-1cm}
\end{wraptable}

We compare the optimal configurations for several compute budgets in Table~\ref{tab:opt_n_d_table}.

Both from comparing the IsoFLOP slices (Figure~\ref{fig:isoflops}) and the values listed in the table, we can see that the compute-optimal configuration for a given compute budget clearly depends on $E$, with MoE models requiring comparatively larger datasets and correspondingly fewer active parameters. 

\begin{wrapfigure}{r}{0.5\textwidth}
\vspace{-0.5cm}
\begin{minipage}{\linewidth}
\begin{mdframed}[
  backgroundcolor=gray!20,
  linecolor=black,
  linewidth=0pt,
  roundcorner=5pt,
  innertopmargin=1em,
  innerbottommargin=1em,
  nobreak=true
]
\textbf{Finding 2.} \textbf{More experts $\boldsymbol{\rightarrow}$ better performance.}
\\For a given compute budget, increasing the number of experts always improves performance, provided the size of the model and the number of training tokens are adjusted (Figure~\ref{fig:isoflops} (a)).
\end{mdframed}
\end{minipage}
\vspace{-0.5cm}
\end{wrapfigure}

\textbf{Mixture of Experts is Compute Optimal.} We now compare the performance across various numbers of experts, with the respective values of tokens and active parameters optimized. As illustrated in Figure~\ref{fig:isoflops}, we observe significant compute savings for MoE models compared to dense models, with a larger number of experts providing more pronounced benefits.

The higher efficiency of MoE in terms of training compute comes at the price of increased memory requirements. However, somewhat surprisingly, we find that MoE models can outperform dense models \textit{of the same size} trained with the same amount of training compute—a result we describe in more detail in the next subsection.

\newpage
\subsection{Model Memory Optimality}
Often, it is insufficient to consider models solely from the perspective of compute optimality, as a compute-optimal model can be impractically large, preventing its deployment on available hardware. 
Additionally, it may only be possible to run a large model with a small batch size due to limited GPU memory, leading to low hardware utilization~\citep{he2022brrrrfromfirstprinciples}. Therefore, it is natural to consider a straightforward extension to the notion of compute optimality, specifically model memory optimality. A model is said to be memory optimal if, among models trained with the same compute budget $F$ and having at most $M$ parameters, it achieves the lowest loss:
\begin{align*}
  &\argmin_{\Nact, D, E} \mathcal{L}(\Nact, D, E) \\
    &\text{s.t. } 6 \Nact D = F, \quad \Ntotal \leq M
\end{align*}
\vspace{-0.5cm}

Note that model memory-matched dense and MoE models differ in the number of active parameters---MoE uses just a fraction of them. Intuitively, it should thus have worse performance. However, given some budget, it can be trained on more tokens, lowering the loss. Our scaling laws suggest that MoE models can be model memory optimal. We validate this claim by training a $1.1$B dense model and a model size and FLOP matched $E=\{2, 4\}$ counterparts (Figure~\ref{fig:thesis_proof}). Significantly, the MoE models attain lower loss even if the dense model is overtrained (i.e., after passing its compute-optimal token count).  

\begin{wrapfigure}{r}{0.5\textwidth}
\vspace{-0.5cm}
\begin{minipage}{\linewidth}
\begin{mdframed}[
  backgroundcolor=gray!20,
  linecolor=black,
  linewidth=0pt,
  roundcorner=5pt,
  innertopmargin=1em,
  innerbottommargin=1em,
  nobreak=true,
]
\textbf{Finding 3.} \textbf{MoE can also be \textit{memory} optimal.} \\ 
A total-parameter-matched MoE model can outperform a dense model trained with the same compute budget (Figure~\ref{fig:thesis_proof}). Moreover, such an MoE model is more compute \textit{and} memory efficient at inference.
\end{mdframed}
\end{minipage}
\vspace{-1cm}
\end{wrapfigure}

\subsection{Total Memory Optimality}
During autoregressive generation, a decoder-only model processes a single token while storing activations (keys and values) for previous tokens in the KV~cache. In the case of multi-head attention, its size equals $2T \times N_{\text{blocks}} \times d_{\text{model}}$, where $T$ is the number of tokens in the cache (possibly within multiple sequences in the batch). Including the cache size yields the optimization criterion:
\begin{align*}
  &\argmin_{\Nact, D, E} \mathcal{L}(\Nact, D, E) \\
  &\text{s.t. } 6\Nact D = F, \quad \Ntotal + 2T N_{\text{blocks}} \dm \leq M
\end{align*}
\vspace{-0.5cm}

For practical values of $T$, a fair comparison of memory requirements should include the size of the KV~cache in addition to the model size. Figure~\ref{fig:optimal_regions}~(b) presents the optimal models for a given compute and varying memory constraints when the size of the KV~cache is included. Importantly, MoE models compare more favorably to dense models in this graph, and as $T$ increases, they outperform dense models at increasingly smaller model sizes. In Figure~\ref{fig:thesis_proof}~(b), the $E=\{2, 4\}$ models employ a smaller KV~cache, which means that if memory is constrained, the MoE model can store longer contexts or work with a larger batch size than the dense model.

\subsection{Inference Optimality}

Large models, while capable, may also be too costly to operate due to their high computational demands. To account for this drawback, we can further assume that a model will process a number of tokens, $D_{\text{inf}}$, over its lifetime and find the best model whose demands do not exceed a predefined joint training and inference budget:
\begin{align*}
  &\argmin_{\Nact, D, E} \mathcal{L}(\Nact, D, E) \\
    &\text{s.t. } 6 \Nact D + 2 \Nact D_{\text{inf}} = F.
\end{align*}
\vspace{-0.5cm}

\begin{figure*}[h]
    \centering
    \subfigure[]{
        \includegraphics[width=0.31\textwidth]{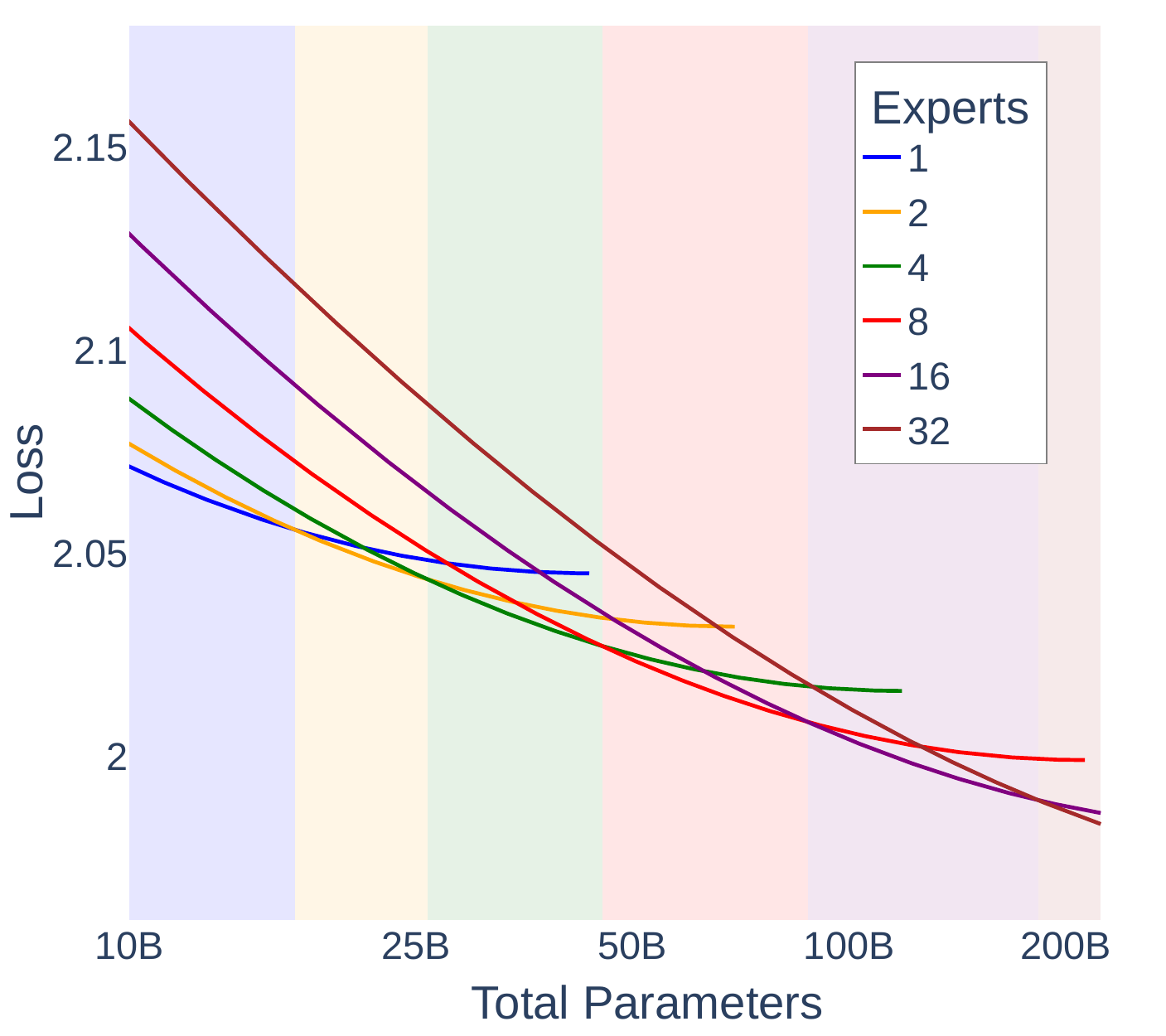}
    }
    \subfigure[]{
        \includegraphics[width=0.31\textwidth]{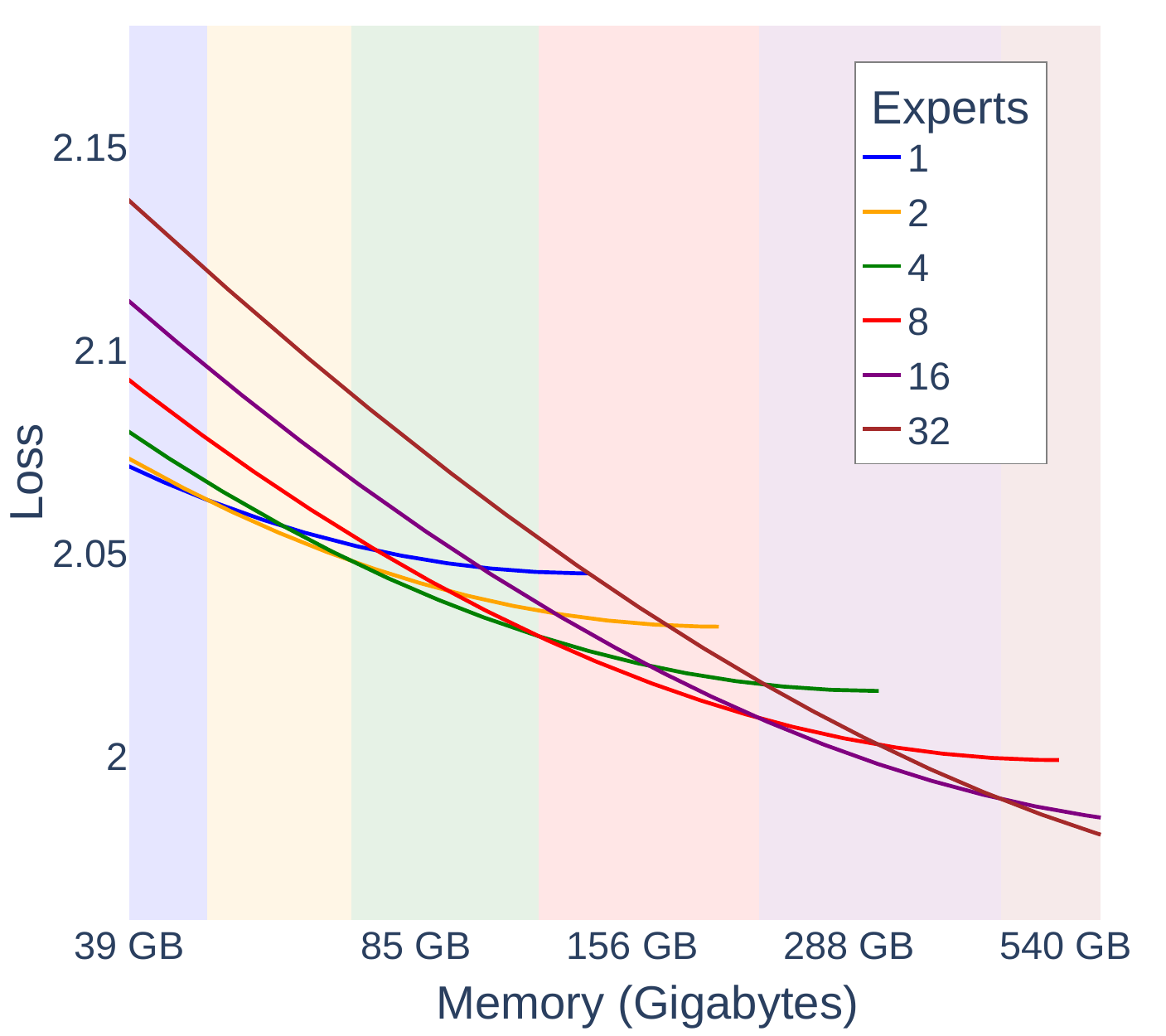}
    }
    \subfigure[]{
        \includegraphics[width=0.31\textwidth]{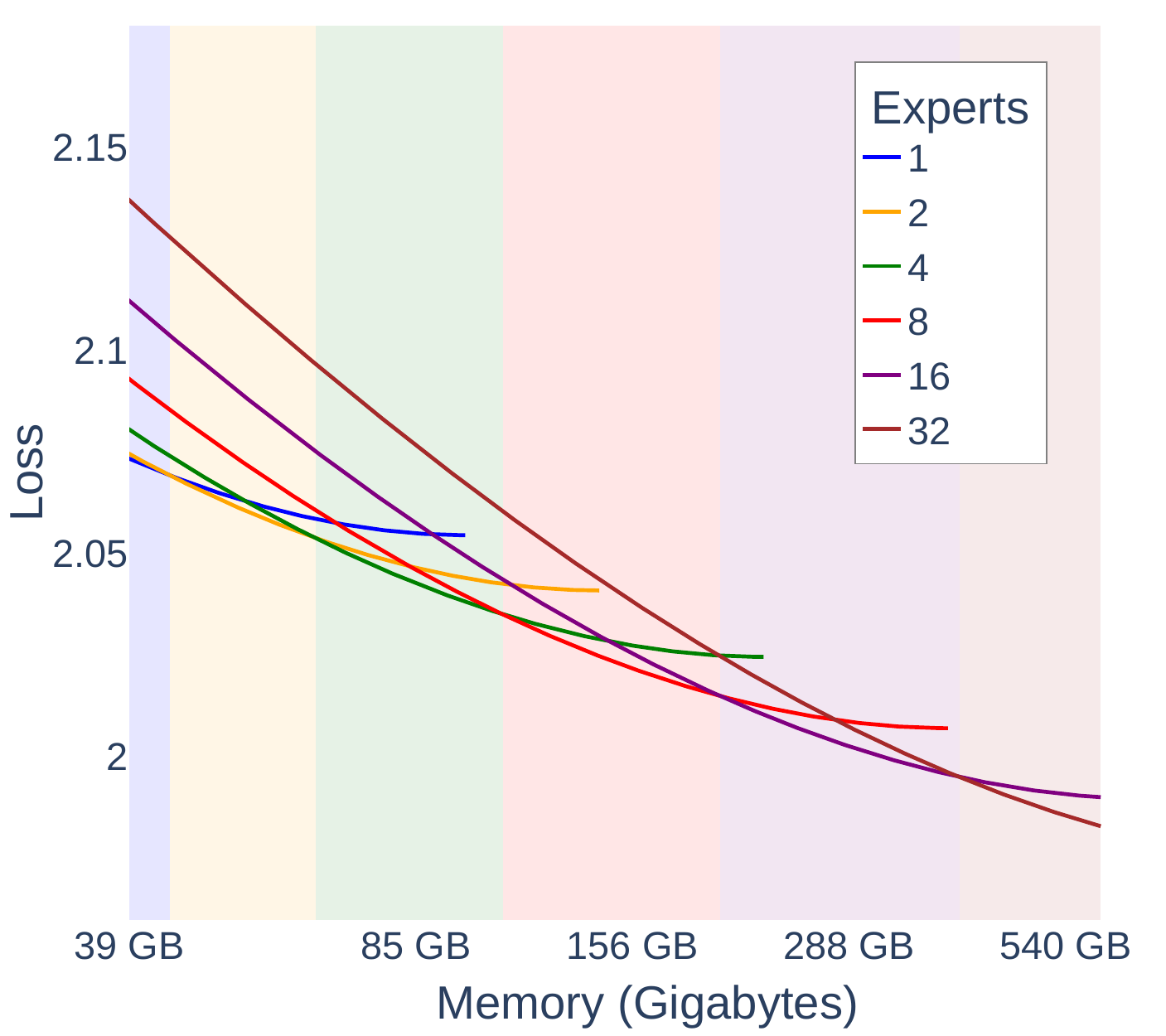}
    }
    \caption{Predicted loss for various numbers of experts at a FLOPs budget $F= 5 \times 10^{22}$. The x-axis represents the size of the model in terms of the number of parameters (a) or the total memory budget for both model parameters and KV~cache for $8192$ tokens (b, c). Shaded areas indicate the optimal number of experts for the corresponding parameter or memory budget. (c) In addition to the KV~cache, the inference cost on $100$B tokens is included in the FLOPs budget of $F= 5 \times 10^{22}$.}
    
  \label{fig:optimal_regions}
\end{figure*}

Figure~\ref{fig:optimal_regions} (c) presents the optimal models for a given compute and varying memory constraints if a joint budget needs to accommodate both training and inference demands. We find that, in this scenario, MoE models outperform dense models at smaller scales than in simple compute optimality due to reduced inference FLOPs. The $E=2$ and $E=4$ models shown in Figure~\ref{fig:thesis_proof} use 36\% and 61\% less FLOPs per token, respectively, than their dense counterparts.

\newpage

\subsection{Summary}

\begin{figure*}[h]
    \begin{center}
        \vspace{0.5cm}
      \includegraphics[width=1\textwidth]{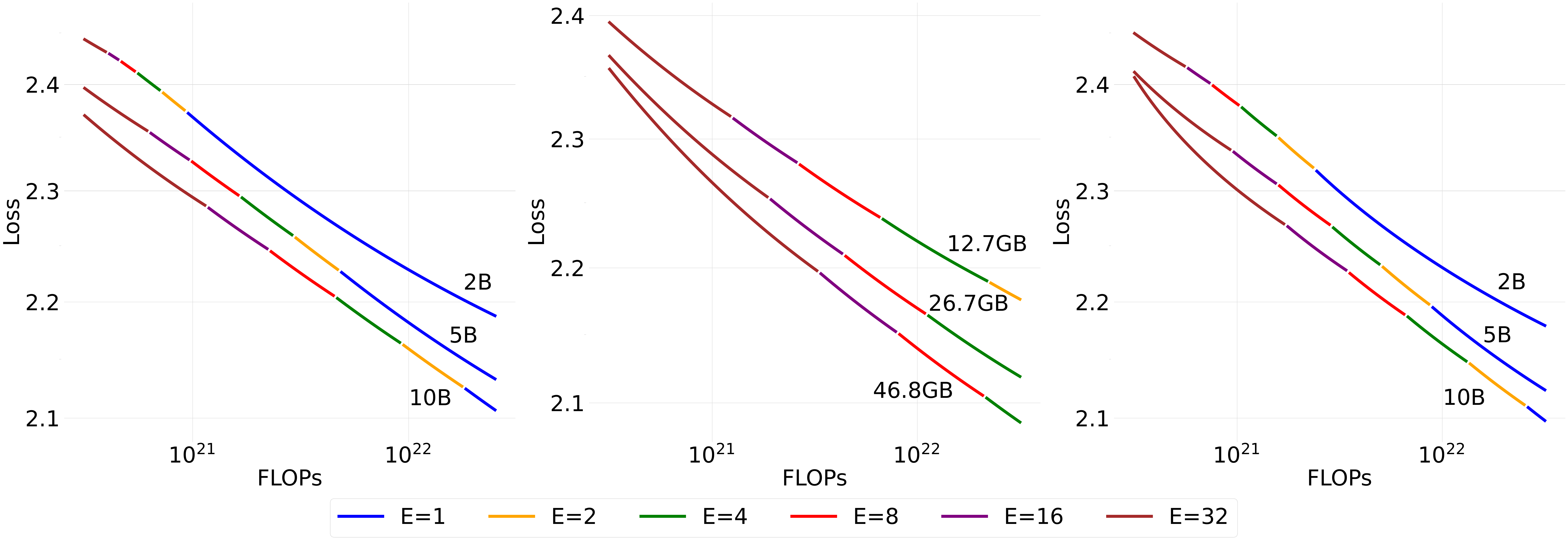}
    \end{center}
    \caption{Investigation of the optimal number of experts for three different model sizes: $2$B, $5$B, and $10$B; and in three different scenarios from left to right: simply measuring the model size, including the size of a KV-cache with 32k tokens, and including the inference cost of processing 100B tokens. Note that in the second graph, the memory constraint corresponds to the memory requirements of dense models with sizes $2$B, $5$B, and $10$B, including the KV~cache, while utilizing bfloat16 for both parameters and activations.
    }
    \label{fig:optimal_l_f}
\end{figure*}

The concepts of inference optimality and total memory optimality can naturally be combined. Figure~\ref{fig:optimal_regions}(c) presents a comparison between different numbers of experts, where the KV~cache is included in the model's memory requirements and the compute budget is shared between training and inference. Finally, Figure~\ref{fig:optimal_l_f} and Table~\ref{tab:optimal_e} investigate the optimal $E$ for a sample of model sizes, while including the KV~cache and considering the inference cost.

\begin{wraptable}{r}{0.5\textwidth}
\begin{center}
\begin{tabular}{cccc}
\toprule
 & 24GB & 80GB & 640GB \\
\midrule
$10^{21}$ & \textcolor{e16}{16} & \textcolor{e32}{$\geq 32$} & \textcolor{e32}{$\geq 32$} \\
$10^{22}$ & \textcolor{e4}{4} & \textcolor{e16}{16} & \textcolor{e32}{$\geq 32$} \\
$10^{23}$ & \textcolor{e1}{1} & \textcolor{e8}{8} & \textcolor{e32}{$\geq 32$} \\
$10^{24}$ & \textcolor{e1}{1} & \textcolor{e1}{1} & \textcolor{e16}{16} \\
\bottomrule
\end{tabular}

\caption{\small Optimal $E$ for different training budgets and three typical memory constraints, corresponding to an RTX4090 GPU, an H100 GPU, and an 8xH100 GPU node. We assume 16k tokens in the KV~cache and bfloat16 for storing model weights and activations. 
}
\label{tab:optimal_e}
\end{center}
\vspace{-2.0cm}
\end{wraptable}

For practitioners, as a simplification of our analysis, we propose a general rule of thumb:

\begin{minipage}{1.0\linewidth}
\begin{mdframed}[
  backgroundcolor=blue!10,
  linecolor=blue!50!black,
  linewidth=0pt,
  innerleftmargin=10pt, innerrightmargin=10pt,
  innertopmargin=1em, innerbottommargin=1em,
  roundcorner=3pt,
]
\textbf{Rule of Thumb.}
For a fixed total parameter count, an MoE model with $E \leq 8$ experts outperforms a compute-optimal dense model if trained on $E$ times more tokens while maintaining the same memory footprint.
\end{mdframed}
\end{minipage}

For instance, a compute-optimal $1.1$B model trained for $8$B
tokens will have worse loss than either a $2$-expert, $1.1$B total parameters MoE model trained on $16$B tokens or a $4$-expert, $1.1$B total parameters MoE model trained on $32$B tokens. At the same time the MoE models will require fewer FLOPs per token during inference.

Note that in the scenario described by the rule of thumb, compute-matched MoE will generally have less than $E$-times larger dataset and will still surpass dense model (as in Figure \ref{fig:thesis_proof} (b)), but we wanted to keep this rule simple and conservative. Furthermore, while the rule may plausibly apply with $E > 8$, we prefer to conservatively limit it to $E \leq 8$ due to the uncertainty of predicting the loss of highly overtrained models (i.e., with a large token-to-parameter ratio). A detailed comparison can be found in Figure~\ref{plot:com_and_mem_matched}, illustrating a stronger result where memory- and compute-matched MoE outperform compute-optimal dense models across scales.

It is important to recognize that such scaling depends on access to large datasets---a concern frequently raised in the context of scaling LLMs. While many leading organizations have demonstrated that data limitations can be overcome, the availability of large-scale datasets varies by organization and domain, particularly outside of NLP. Whether NLP datasets are effectively unlimited remains an open question beyond the scope of this work.

\newpage

\newpage

\vspace{0.3cm}
\section{Fitting the Scaling Law}\label{sec: fitting_scaling_law}

In this section, we present the details of the experiments and the procedure for fitting the scaling law parameters, see Table \ref{tab:scaling_law_params} in the Appendix. These results are based on extensive large-scale empirical evidence, including over $280$ models with up to $5$B parameters, trained on a variety of compute budgets. For a comprehensive list of experiments, see Appendix~\ref{sec:full_experiments}.

\vspace{0.2cm}
\subsection{Model Hyperparameters}

The selection of hyperparameters and training details is crucial for ensuring the robustness of scaling laws~\citep{porian2025resolvingdiscrepanciescomputeoptimalscaling,pearce2024reconcilingkaplanchinchillascaling}. In our work, we employ a set of best practices and modern design choices, aiming to provide accurate predictions applicable to real-life practice.

All models used in this study are decoder-only Transformers trained on the highly 
filtered FineWeb-Edu~\citep{penedo2024finewebdatasetsdecantingweb}. We use a Transformer model with Switch~\citep{fedus2022switch} layers, using standard values of router z-loss of $0.001$ and load balancing loss of $0.01$. The GPT-2 tokenizer~\citep{radford2018improving} is employed. For better stability, weight initialization follows a truncated normal distribution with a reduced scale of $0.1$, as suggested by \cite{fedus2022switch}. Mixed precision training is used, with the attention mechanism, RoPE position embeddings \cite{su2023roformerenhancedtransformerrotary} and router always maintained at high precision. The models use the SwiGLU activation ~\citep{shazeer2020gluvariantsimprovetransformer} with hidden size equal to $3\text{d}_\text{model}$  and activate one expert per token (unless the token is dropped due to limited capacity). For evaluation, we increase the capacity factor to ensure dropless processing of the tokens.

\subsubsection{Batch Size Ramp-Up}
Performance of a deep learning optimization procedure can suffer as a result of using an exceedingly large batch size~\citep{mccandlish2018empiricalmodellargebatchtraining}. To mitigate this potential issue, especially early in the training, we employ batch-size ramp-up. Similar strategies are used in contemporary LLM training runs~\citep{rae2022scaling, dubey2024llama}. We increase the batch size from $64$K to $128$K after $0.5$B training tokens and further to $256$K after $1$B training tokens. Instead of utilizing noise scale as a critical batch size predictor~\citep{mccandlish2018empiricalmodellargebatchtraining} we opted for a straightforward grid to directly predict a transition point beyond which increasing batch size does not impair performance.

\newpage
\subsubsection{Learning Rate Scaling}

\citet{kaplan2020scaling} have shown that scaling laws for hyperparameters can be used to adjust them according to the size of the model in the case of dense Transformers. For MoE models, we find the literature inconclusive---while some \citep{dai2024deepseekmoe}
pretrain MoEs with a lower learning rate than corresponding dense models, others \citep{zoph2022st} report better performance when fine-tuning MoEs with higher learning rates. To address this discrepancy, we derive a scaling law for the peak learning rate for MoE based on the number of active non-embedding parameters $\Nane$ and the number of experts $E$:
\vspace{-0.1cm}
\begin{equation}\label{eq:lr_scaling}
    LR(\Nane, E) = \exp(8.39 - 0.81 \log(\Nane) -0.25 \log(E)),
\end{equation}
\vspace{-0.5cm}
\begin{wrapfigure}{r}{0.5\textwidth}
\vspace{-0.2cm}
\begin{minipage}{\linewidth}
\begin{mdframed}[
  backgroundcolor=gray!20,
  linecolor=black,
  linewidth=0pt,
  roundcorner=5pt,
  innertopmargin=1em,
  innerbottommargin=1em,
  nobreak=true
]
\textbf{Finding 4.} \textbf{More experts $\boldsymbol{\rightarrow}$ lower learning rate.} \\ 
Increasing the number of experts in MoE model should be accompanied by lowering the learning rate accordingly (Figure~\ref{plot:lr_scaling_with_e} in the Appendix).
\end{mdframed}
\end{minipage}
\vspace{-0.2cm}
\end{wrapfigure}
and use this equation to set the learning rate in our main scaling laws experiments. We fit the coefficients of the equation using the least squares method, minimizing the error between the prediction and the optimal learning rate from the experiment grid. In contrast to \citet{kaplan2020scaling}, we use a linear transformation of the parameter count to predict the logarithm of the learning rate, instead of directly predicting the learning rate. This approach allows us to avoid the breakdown of the formula above $10^{10}$ parameters, as mentioned in their work, where the predicted learning rate becomes negative. This phenomenon is independent of the actual fit and is simply a property of the formula used. Besides being well-defined in the extrapolation, we argue that optimal learning rates visibly follow this logarithmic trend, as seen in Figure~\ref{plot:lr_scaling_with_e} in the Appendix.

The second difference between our formula and the one by \citet{kaplan2020scaling} is the incorporation of the number of experts, allowing us to model the optimal behavior of this hyperparameter across dense models and different MoEs. 
This is an important detail that allows unbiased comparison among different models and ensures each one is optimally tuned. Furthermore, it allows us to answer the question of whether MoE should be trained with a lower or higher learning rate. While our formula accommodates both scenarios, we can clearly see in Figure~\ref{plot:lr_scaling_with_e} in the Appendix that increasing $E$ requires lower learning rates, resulting in a negative value for the coefficient. Moreover, we verify this thesis by tuning the fit on $E=1$ and $E=8$, and validating it on interpolation at $E=4$ and extrapolation at $E=32$. In both instances, the validation predicts the optimal learning rate for the model configuration or a value with nearly the same performance.

In Figure~\ref{plot:lr_scaling_no_e} in the Appendix, we perform an ablation of this additional power law on $E$ by repeating our entire fitting procedure without the $E$ component. This shows, especially with extrapolation on $E=32$, that dependence on $E$ is crucial, and its omission can impair the performance of MoEs. Further details about our scaling rule for learning rates can be found in the plots in Appendix~\ref{app:lr_scaling}.

\subsubsection{Learning Rate Schedule}
\citet{hägele2024scalinglawscomputeoptimaltraining} suggests that a trapezoidal learning rate schedule can yield similar performance to other established methods, such as the cosine schedule. Additionally, it provides a valuable advantage when varying training duration, as intermediate checkpoints can be reused. With a cosine schedule, intermediate checkpoints introduce bias into the fit, according to the analysis of \citet{kaplan2020scaling} by \citet{hoffmann2022training}.
We employ a constant learning rate schedule with a linear warmup over the initial $130$M tokens and with a linear decay from the peak learning rate to $0$ over the final $20\%$ of tokens. For each model size, longer runs reuse intermediate checkpoints from the shorter ones.

\subsection{Optimization of Formula Coefficients} \label{section:fitting}
Following \citet{hoffmann2022training}, we use the LBFGS algorithm to optimize the coefficients of Equation~\ref{eq:scaling_law_final}. See Appendix~\ref{app:fit_details} for details. We observe a good fit with $\texttt{RMSE}_{v}=0.0039$ on a held-out set of our $30$ runs with the lowest loss, and $\texttt{RMSE}_t=0.0062$ on the training dataset. To further verify our formula, we train separate Chinchilla scaling laws (Equation~\ref{eq:scaling_law_chinchilla}) for different $E$ using the same hyperparameters and the corresponding subset of the initializations grid. This approach serves as a lower bound for the loss of our joint formula on the training dataset, as it can emulate its coefficients; however, it is more prone to overfitting because effectively more parameters are utilized. Using this approach, we obtain a lower error on the training dataset of $\texttt{RMSE}_{t}^{\text{sep}}=0.0059$ and marginally higher on the validation $\texttt{RMSE}_{v}^{\text{sep}}=0.0041$. We believe this is a strong confirmation that our joint formula is actually describing how variable $E$ influences training.

\begin{figure*}[h]
    \begin{center}
      \includegraphics[width=.33\textwidth]{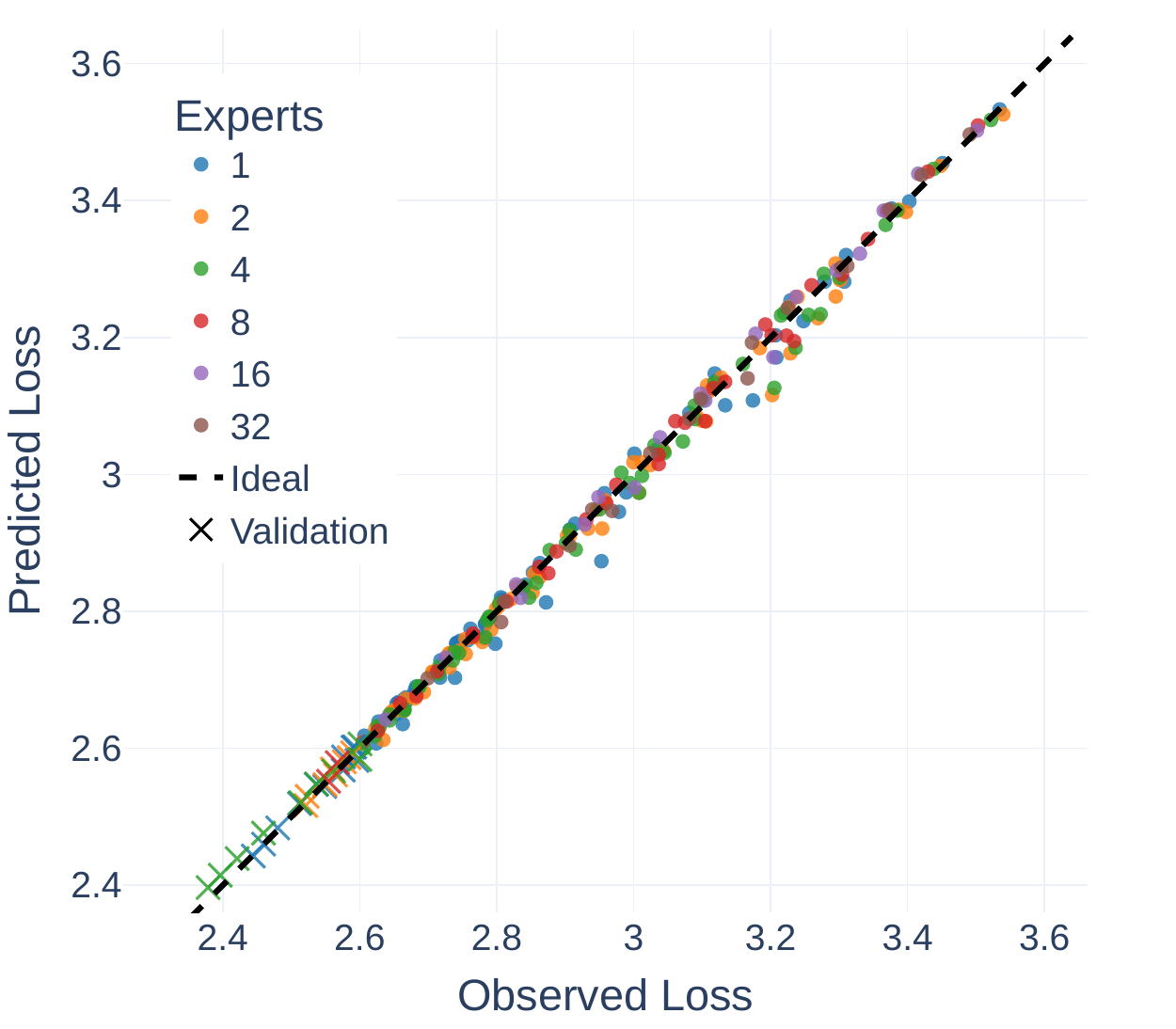}
      \includegraphics[width=.33\textwidth]{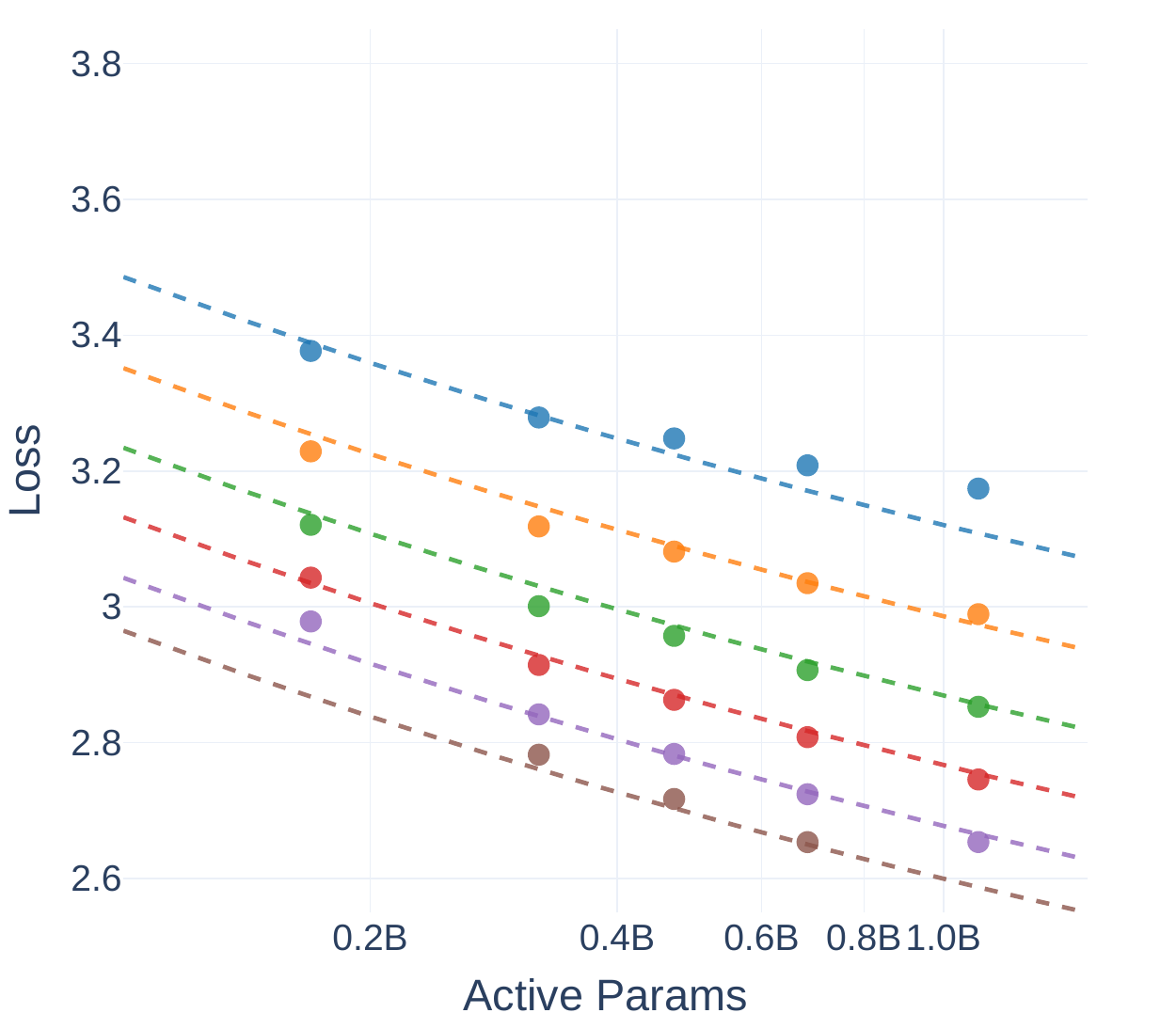}
      \includegraphics[width=.33\textwidth]{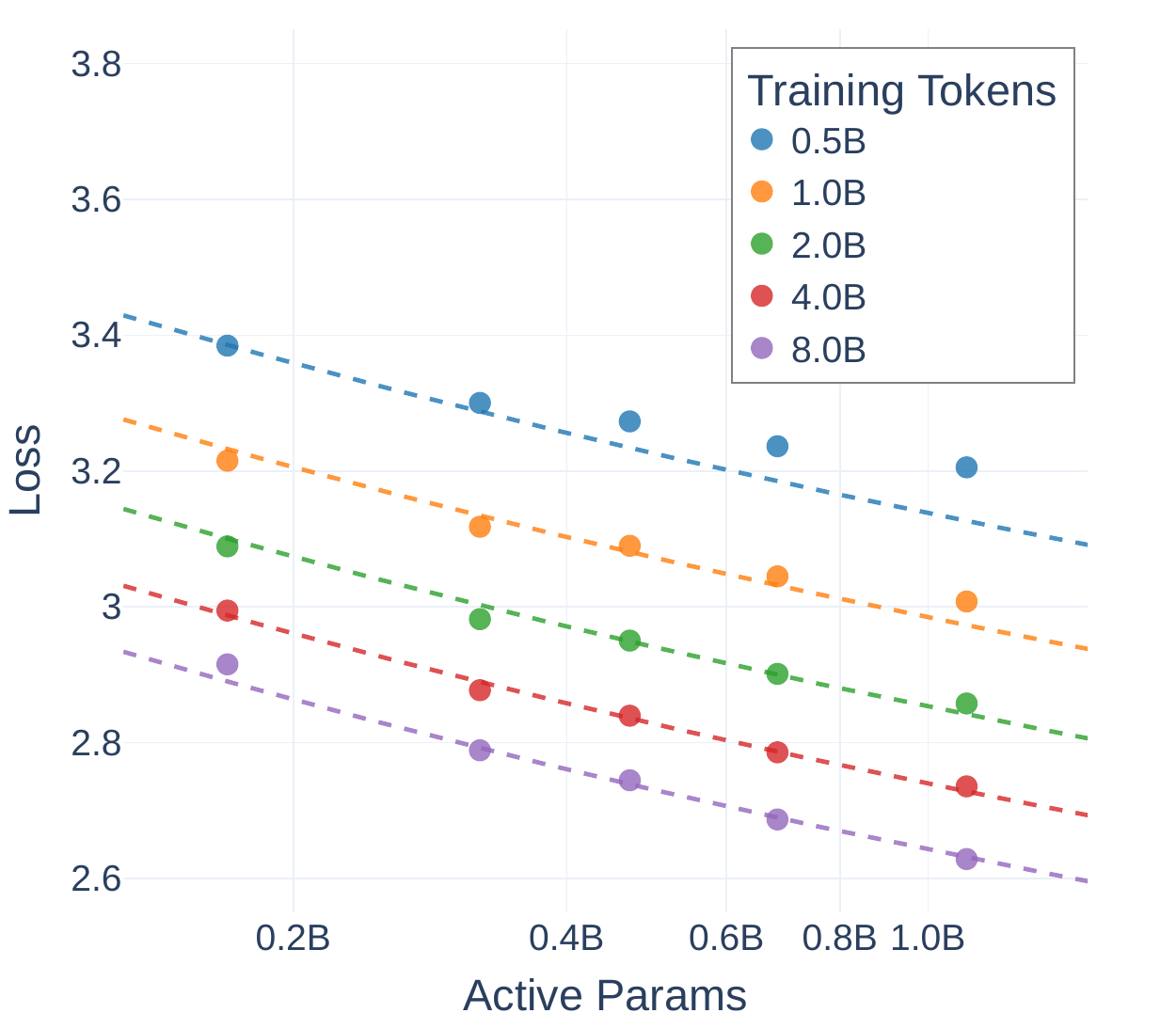} \\
\textbf{(a) \hspace{150px} (b) \hspace{150px} (c)}      
    \end{center}
    \caption{\textbf{(a)} Quality of the fit. The maximum absolute error on the held-out extrapolation set is $0.018$. \textbf{(b)} Predicted loss compared to observed loss for $E=1$. \textbf{(c)} Predicted loss (dashed line) compared to observed loss for $E=4$. We can see that on the training dataset, the error increases in an undertrained setting ($D/N<1$ --- more tokens than parameters). However, this scenario is never practical from our perspective.}
    \label{fig:kropki_kreski}
\end{figure*}

In Figure~\ref{fig:kropki_kreski}, we visually verify the extrapolation of the joint fit. Prediction errors are categorized by different numbers of experts, highlighting that our joint formula is not biased for any specific $E$.

\newpage

\section{Limitations and Future Work}

In our work, we focus on the standard MoE variant, where the size of each expert matches the size of the feed-forward layer in the corresponding dense model. Some recent findings~\citep{dai2024deepseekmoe, ludziejewskiscaling, muennighoff2024olmoeopenmixtureofexpertslanguage, qwen_moe} suggest that fine-grained MoE models are more efficient and may likely enhance the benefits we report for using MoE. Similarly, adopting a dropless MoE~\citep{gale2022megablocksefficientsparsetraining} approach, instead of relying on a capacity factor, could lead to further improvements. We leave the integration of these MoE improvements for future work.

Moreover, our Chinchilla-based optimality analysis utilizes FLOPs, which may not accurately reflect the wall-clock training time of models with different architectures. Although comparing models based on the total number of parameters, rather than active parameters, partially alleviates this issue due to the same memory bottleneck, different implementations and distributed training algorithms are not considered in this work.

We assumed the Chinchilla scaling law (Equation~\ref{eq:scaling_law_chinchilla}) as the basis for our formulas. While this is well-grounded in the literature, this formula is known to have limitations, particularly for extreme token-to-parameter ratios. Similarly, we observed a regression in the goodness of fit for some heavily undertrained or overtrained runs.

\section{Conclusions}

In this work, we derived the joint scaling laws for Mixture of Experts, relating the loss of the model to the number of parameters, the number of training tokens, and the number of experts. By considering both compute and memory constraints, as well as the expected inference workload, we demonstrated that MoE models can outperform dense models even when constrained by memory usage or total parameters, contrary to common assumptions and intuitions that MoE models are more memory-intensive than dense models.

Our analysis reveals how optimal training strategies shift as the number of experts varies. This provides a principled framework for selecting MoE hyperparameters under given constraints, highlighting the trade-offs between memory and compute performance.

\newpage
\section*{Acknowledgments}

We would like to express sincere gratitude to Szymon Antoniak and Piotr Padlewski for their detailed comments and invaluable discussions. We also thank Konrad Staniszewski for his feedback on the draft of this paper.

We gratefully acknowledge the Polish high-performance computing infrastructure PLGrid (HPC Center: ACK Cyfronet AGH) for providing computer facilities and support within computational grant no. PLG/2024/017060. This research was partially supported by the ERC PoC Grant EXALT no. 101082299, the National Science Centre (NCN) Grant no. 2020/37/B/ST6/04179, the National Science Centre (NCN) Preludium Grant no. 2022/45/N/ST6/02222, the "European Lighthouse of AI for Sustainability" - ELIAS grant no. 101120237, and the NCBiR grant POIR.01.01.01-00-0433/20. Part of the experiments utilized computational resources provided by \href{https://writer.com/}{Writer}.

\nocite{langley00}

\bibliography{bib}

\begin{thebibliography}{40}
\providecommand{\natexlab}[1]{#1}
\providecommand{\url}[1]{\texttt{#1}}
\expandafter\ifx\csname urlstyle\endcsname\relax
  \providecommand{\doi}[1]{doi: #1}\else
  \providecommand{\doi}{doi: \begingroup \urlstyle{rm}\Url}\fi

\bibitem[Abnar et~al.(2025)Abnar, Shah, Busbridge, Ali, Susskind, and Thilak]{abnar2025parametersvsflopsscaling}
Abnar, S., Shah, H., Busbridge, D., Ali, A. M.~E., Susskind, J., and Thilak, V.
\newblock Parameters vs flops: Scaling laws for optimal sparsity for mixture-of-experts language models, 2025.
\newblock URL \url{https://arxiv.org/abs/2501.12370}.

\bibitem[Clark et~al.(2022)Clark, de~las Casas, Guy, Mensch, Paganini, Hoffmann, Damoc, Hechtman, Cai, Borgeaud, van~den Driessche, Rutherford, Hennigan, Johnson, Millican, Cassirer, Jones, Buchatskaya, Budden, Sifre, Osindero, Vinyals, Rae, Elsen, Kavukcuoglu, and Simonyan]{clark2022unified}
Clark, A., de~las Casas, D., Guy, A., Mensch, A., Paganini, M., Hoffmann, J., Damoc, B., Hechtman, B., Cai, T., Borgeaud, S., van~den Driessche, G., Rutherford, E., Hennigan, T., Johnson, M., Millican, K., Cassirer, A., Jones, C., Buchatskaya, E., Budden, D., Sifre, L., Osindero, S., Vinyals, O., Rae, J., Elsen, E., Kavukcuoglu, K., and Simonyan, K.
\newblock Unified scaling laws for routed language models, 2022.

\bibitem[Dai et~al.(2024)Dai, Deng, Zhao, Xu, Gao, Chen, Li, Zeng, Yu, Wu, Xie, Li, Huang, Luo, Ruan, Sui, and Liang]{dai2024deepseekmoe}
Dai, D., Deng, C., Zhao, C., Xu, R.~X., Gao, H., Chen, D., Li, J., Zeng, W., Yu, X., Wu, Y., Xie, Z., Li, Y.~K., Huang, P., Luo, F., Ruan, C., Sui, Z., and Liang, W.
\newblock Deepseekmoe: Towards ultimate expert specialization in mixture-of-experts language models, 2024.

\bibitem[DeepSeek-AI et~al.(2025)DeepSeek-AI, Guo, Yang, Zhang, Song, Zhang, Xu, Zhu, Ma, Wang, Bi, Zhang, Yu, Wu, Wu, Gou, Shao, Li, Gao, Liu, Xue, Wang, Wu, Feng, Lu, Zhao, Deng, Zhang, Ruan, Dai, Chen, Ji, Li, Lin, Dai, Luo, Hao, Chen, Li, Zhang, Bao, Xu, Wang, Ding, Xin, Gao, Qu, Li, Guo, Li, Wang, Chen, Yuan, Qiu, Li, Cai, Ni, Liang, Chen, Dong, Hu, Gao, Guan, Huang, Yu, Wang, Zhang, Zhao, Wang, Zhang, Xu, Xia, Zhang, Zhang, Tang, Li, Wang, Li, Tian, Huang, Zhang, Wang, Chen, Du, Ge, Zhang, Pan, Wang, Chen, Jin, Chen, Lu, Zhou, Chen, Ye, Wang, Yu, Zhou, Pan, Li, Zhou, Wu, Ye, Yun, Pei, Sun, Wang, Zeng, Zhao, Liu, Liang, Gao, Yu, Zhang, Xiao, An, Liu, Wang, Chen, Nie, Cheng, Liu, Xie, Liu, Yang, Li, Su, Lin, Li, Jin, Shen, Chen, Sun, Wang, Song, Zhou, Wang, Shan, Li, Wang, Wei, Zhang, Xu, Li, Zhao, Sun, Wang, Yu, Zhang, Shi, Xiong, He, Piao, Wang, Tan, Ma, Liu, Guo, Ou, Wang, Gong, Zou, He, Xiong, Luo, You, Liu, Zhou, Zhu, Xu, Huang, Li, Zheng, Zhu, Ma, Tang, Zha, Yan, Ren, Ren, Sha, Fu, Xu, Xie, Zhang,
  Hao, Ma, Yan, Wu, Gu, Zhu, Liu, Li, Xie, Song, Pan, Huang, Xu, Zhang, and Zhang]{deepseekr1}
DeepSeek-AI, Guo, D., Yang, D., Zhang, H., Song, J., Zhang, R., Xu, R., Zhu, Q., Ma, S., Wang, P., Bi, X., Zhang, X., Yu, X., Wu, Y., Wu, Z.~F., Gou, Z., Shao, Z., Li, Z., Gao, Z., Liu, A., Xue, B., Wang, B., Wu, B., Feng, B., Lu, C., Zhao, C., Deng, C., Zhang, C., Ruan, C., Dai, D., Chen, D., Ji, D., Li, E., Lin, F., Dai, F., Luo, F., Hao, G., Chen, G., Li, G., Zhang, H., Bao, H., Xu, H., Wang, H., Ding, H., Xin, H., Gao, H., Qu, H., Li, H., Guo, J., Li, J., Wang, J., Chen, J., Yuan, J., Qiu, J., Li, J., Cai, J.~L., Ni, J., Liang, J., Chen, J., Dong, K., Hu, K., Gao, K., Guan, K., Huang, K., Yu, K., Wang, L., Zhang, L., Zhao, L., Wang, L., Zhang, L., Xu, L., Xia, L., Zhang, M., Zhang, M., Tang, M., Li, M., Wang, M., Li, M., Tian, N., Huang, P., Zhang, P., Wang, Q., Chen, Q., Du, Q., Ge, R., Zhang, R., Pan, R., Wang, R., Chen, R.~J., Jin, R.~L., Chen, R., Lu, S., Zhou, S., Chen, S., Ye, S., Wang, S., Yu, S., Zhou, S., Pan, S., Li, S.~S., Zhou, S., Wu, S., Ye, S., Yun, T., Pei, T., Sun, T., Wang, T., Zeng, W.,
  Zhao, W., Liu, W., Liang, W., Gao, W., Yu, W., Zhang, W., Xiao, W.~L., An, W., Liu, X., Wang, X., Chen, X., Nie, X., Cheng, X., Liu, X., Xie, X., Liu, X., Yang, X., Li, X., Su, X., Lin, X., Li, X.~Q., Jin, X., Shen, X., Chen, X., Sun, X., Wang, X., Song, X., Zhou, X., Wang, X., Shan, X., Li, Y.~K., Wang, Y.~Q., Wei, Y.~X., Zhang, Y., Xu, Y., Li, Y., Zhao, Y., Sun, Y., Wang, Y., Yu, Y., Zhang, Y., Shi, Y., Xiong, Y., He, Y., Piao, Y., Wang, Y., Tan, Y., Ma, Y., Liu, Y., Guo, Y., Ou, Y., Wang, Y., Gong, Y., Zou, Y., He, Y., Xiong, Y., Luo, Y., You, Y., Liu, Y., Zhou, Y., Zhu, Y.~X., Xu, Y., Huang, Y., Li, Y., Zheng, Y., Zhu, Y., Ma, Y., Tang, Y., Zha, Y., Yan, Y., Ren, Z.~Z., Ren, Z., Sha, Z., Fu, Z., Xu, Z., Xie, Z., Zhang, Z., Hao, Z., Ma, Z., Yan, Z., Wu, Z., Gu, Z., Zhu, Z., Liu, Z., Li, Z., Xie, Z., Song, Z., Pan, Z., Huang, Z., Xu, Z., Zhang, Z., and Zhang, Z.
\newblock Deepseek-r1: Incentivizing reasoning capability in llms via reinforcement learning, 2025.
\newblock URL \url{https://arxiv.org/abs/2501.12948}.

\bibitem[Du et~al.(2022)Du, Huang, Dai, Tong, Lepikhin, Xu, Krikun, Zhou, Yu, Firat, Zoph, Fedus, Bosma, Zhou, Wang, Wang, Webster, Pellat, Robinson, Meier-Hellstern, Duke, Dixon, Zhang, Le, Wu, Chen, and Cui]{du2022glam}
Du, N., Huang, Y., Dai, A.~M., Tong, S., Lepikhin, D., Xu, Y., Krikun, M., Zhou, Y., Yu, A.~W., Firat, O., Zoph, B., Fedus, L., Bosma, M., Zhou, Z., Wang, T., Wang, Y.~E., Webster, K., Pellat, M., Robinson, K., Meier-Hellstern, K., Duke, T., Dixon, L., Zhang, K., Le, Q.~V., Wu, Y., Chen, Z., and Cui, C.
\newblock Glam: Efficient scaling of language models with mixture-of-experts, 2022.

\bibitem[Dubey et~al.(2024)Dubey, Jauhri, Pandey, Kadian, Al-Dahle, Letman, Mathur, Schelten, Yang, Fan, et~al.]{dubey2024llama}
Dubey, A., Jauhri, A., Pandey, A., Kadian, A., Al-Dahle, A., Letman, A., Mathur, A., Schelten, A., Yang, A., Fan, A., et~al.
\newblock The llama 3 herd of models.
\newblock \emph{arXiv preprint arXiv:2407.21783}, 2024.

\bibitem[Fedus et~al.(2022)Fedus, Zoph, and Shazeer]{fedus2022switch}
Fedus, W., Zoph, B., and Shazeer, N.
\newblock Switch transformers: Scaling to trillion parameter models with simple and efficient sparsity, 2022.

\bibitem[Frantar et~al.(2023)Frantar, Riquelme, Houlsby, Alistarh, and Evci]{frantar2023scaling}
Frantar, E., Riquelme, C., Houlsby, N., Alistarh, D., and Evci, U.
\newblock Scaling laws for sparsely-connected foundation models, 2023.

\bibitem[Gale et~al.(2022)Gale, Narayanan, Young, and Zaharia]{gale2022megablocksefficientsparsetraining}
Gale, T., Narayanan, D., Young, C., and Zaharia, M.
\newblock Megablocks: Efficient sparse training with mixture-of-experts, 2022.
\newblock URL \url{https://arxiv.org/abs/2211.15841}.

\bibitem[Ghorbani et~al.(2021)Ghorbani, Firat, Freitag, Bapna, Krikun, Garcia, Chelba, and Cherry]{ghorbani2021scaling}
Ghorbani, B., Firat, O., Freitag, M., Bapna, A., Krikun, M., Garcia, X., Chelba, C., and Cherry, C.
\newblock Scaling laws for neural machine translation, 2021.

\bibitem[He(2022)]{he2022brrrrfromfirstprinciples}
He, H.
\newblock Making deep learning go brrrr from first principles.
\newblock 2022.
\newblock URL \url{https://horace.io/brrr_intro.html}.

\bibitem[Henighan et~al.(2020)Henighan, Kaplan, Katz, Chen, Hesse, Jackson, Jun, Brown, Dhariwal, Gray, Hallacy, Mann, Radford, Ramesh, Ryder, Ziegler, Schulman, Amodei, and McCandlish]{henighan2020scaling}
Henighan, T., Kaplan, J., Katz, M., Chen, M., Hesse, C., Jackson, J., Jun, H., Brown, T.~B., Dhariwal, P., Gray, S., Hallacy, C., Mann, B., Radford, A., Ramesh, A., Ryder, N., Ziegler, D.~M., Schulman, J., Amodei, D., and McCandlish, S.
\newblock Scaling laws for autoregressive generative modeling, 2020.

\bibitem[Hestness et~al.(2017)Hestness, Narang, Ardalani, Diamos, Jun, Kianinejad, Patwary, Yang, and Zhou]{hestness2017deep}
Hestness, J., Narang, S., Ardalani, N., Diamos, G., Jun, H., Kianinejad, H., Patwary, M. M.~A., Yang, Y., and Zhou, Y.
\newblock Deep learning scaling is predictable, empirically, 2017.

\bibitem[Hochreiter \& Schmidhuber(1997)Hochreiter and Schmidhuber]{hochreiter1997long}
Hochreiter, S. and Schmidhuber, J.
\newblock Long short-term memory.
\newblock \emph{Neural computation}, 9\penalty0 (8):\penalty0 1735--1780, 1997.

\bibitem[Hoffmann et~al.(2022)Hoffmann, Borgeaud, Mensch, Buchatskaya, Cai, Rutherford, de~Las~Casas, Hendricks, Welbl, Clark, Hennigan, Noland, Millican, van~den Driessche, Damoc, Guy, Osindero, Simonyan, Elsen, Rae, Vinyals, and Sifre]{hoffmann2022training}
Hoffmann, J., Borgeaud, S., Mensch, A., Buchatskaya, E., Cai, T., Rutherford, E., de~Las~Casas, D., Hendricks, L.~A., Welbl, J., Clark, A., Hennigan, T., Noland, E., Millican, K., van~den Driessche, G., Damoc, B., Guy, A., Osindero, S., Simonyan, K., Elsen, E., Rae, J.~W., Vinyals, O., and Sifre, L.
\newblock Training compute-optimal large language models, 2022.

\bibitem[Hägele et~al.(2024)Hägele, Bakouch, Kosson, Allal, Werra, and Jaggi]{hägele2024scalinglawscomputeoptimaltraining}
Hägele, A., Bakouch, E., Kosson, A., Allal, L.~B., Werra, L.~V., and Jaggi, M.
\newblock Scaling laws and compute-optimal training beyond fixed training durations, 2024.
\newblock URL \url{https://arxiv.org/abs/2405.18392}.

\bibitem[Jacobs et~al.(1991)Jacobs, Jordan, Nowlan, and Hinton]{moe1991}
Jacobs, R.~A., Jordan, M.~I., Nowlan, S.~J., and Hinton, G.~E.
\newblock Adaptive mixtures of local experts.
\newblock \emph{Neural Computation}, 3\penalty0 (1):\penalty0 79--87, 1991.
\newblock \doi{10.1162/neco.1991.3.1.79}.

\bibitem[Jiang et~al.(2024)Jiang, Sablayrolles, Roux, Mensch, Savary, Bamford, Chaplot, de~las Casas, Hanna, Bressand, Lengyel, Bour, Lample, Lavaud, Saulnier, Lachaux, Stock, Subramanian, Yang, Antoniak, Scao, Gervet, Lavril, Wang, Lacroix, and Sayed]{jiang2024mixtral}
Jiang, A.~Q., Sablayrolles, A., Roux, A., Mensch, A., Savary, B., Bamford, C., Chaplot, D.~S., de~las Casas, D., Hanna, E.~B., Bressand, F., Lengyel, G., Bour, G., Lample, G., Lavaud, L.~R., Saulnier, L., Lachaux, M.-A., Stock, P., Subramanian, S., Yang, S., Antoniak, S., Scao, T.~L., Gervet, T., Lavril, T., Wang, T., Lacroix, T., and Sayed, W.~E.
\newblock Mixtral of experts, 2024.

\bibitem[Kaplan et~al.(2020)Kaplan, McCandlish, Henighan, Brown, Chess, Child, Gray, Radford, Wu, and Amodei]{kaplan2020scaling}
Kaplan, J., McCandlish, S., Henighan, T., Brown, T.~B., Chess, B., Child, R., Gray, S., Radford, A., Wu, J., and Amodei, D.
\newblock Scaling laws for neural language models, 2020.

\bibitem[Kumar et~al.(2024)Kumar, Ankner, Spector, Bordelon, Muennighoff, Paul, Pehlevan, Ré, and Raghunathan]{kumar2024scalinglawsprecision}
Kumar, T., Ankner, Z., Spector, B.~F., Bordelon, B., Muennighoff, N., Paul, M., Pehlevan, C., Ré, C., and Raghunathan, A.
\newblock Scaling laws for precision, 2024.
\newblock URL \url{https://arxiv.org/abs/2411.04330}.

\bibitem[Langley(2000)]{langley00}
Langley, P.
\newblock Crafting papers on machine learning.
\newblock In Langley, P. (ed.), \emph{Proceedings of the 17th International Conference on Machine Learning (ICML 2000)}, pp.\  1207--1216, Stanford, CA, 2000. Morgan Kaufmann.

\bibitem[Lepikhin et~al.(2020)Lepikhin, Lee, Xu, Chen, Firat, Huang, Krikun, Shazeer, and Chen]{lepikhin2020gshard}
Lepikhin, D., Lee, H., Xu, Y., Chen, D., Firat, O., Huang, Y., Krikun, M., Shazeer, N., and Chen, Z.
\newblock Gshard: Scaling giant models with conditional computation and automatic sharding, 2020.

\bibitem[Ludziejewski et~al.(2024)Ludziejewski, Krajewski, Adamczewski, Pi\'{o}ro, Krutul, Antoniak, Ciebiera, Kr\'{o}l, Odrzyg\'{o}\'{z}d\'{z}, Sankowski, Cygan, and Jaszczur]{ludziejewskiscaling}
Ludziejewski, J., Krajewski, J., Adamczewski, K., Pi\'{o}ro, M., Krutul, M., Antoniak, S., Ciebiera, K., Kr\'{o}l, K., Odrzyg\'{o}\'{z}d\'{z}, T., Sankowski, P., Cygan, M., and Jaszczur, S.
\newblock Scaling laws for fine-grained mixture of experts.
\newblock In Salakhutdinov, R., Kolter, Z., Heller, K., Weller, A., Oliver, N., Scarlett, J., and Berkenkamp, F. (eds.), \emph{Proceedings of the 41st International Conference on Machine Learning}, volume 235 of \emph{Proceedings of Machine Learning Research}, pp.\  33270--33288. PMLR, 21--27 Jul 2024.
\newblock URL \url{https://proceedings.mlr.press/v235/ludziejewski24a.html}.

\bibitem[McCandlish et~al.(2018)McCandlish, Kaplan, Amodei, and Team]{mccandlish2018empiricalmodellargebatchtraining}
McCandlish, S., Kaplan, J., Amodei, D., and Team, O.~D.
\newblock An empirical model of large-batch training, 2018.
\newblock URL \url{https://arxiv.org/abs/1812.06162}.

\bibitem[Muennighoff et~al.(2024)Muennighoff, Soldaini, Groeneveld, Lo, Morrison, Min, Shi, Walsh, Tafjord, Lambert, Gu, Arora, Bhagia, Schwenk, Wadden, Wettig, Hui, Dettmers, Kiela, Farhadi, Smith, Koh, Singh, and Hajishirzi]{muennighoff2024olmoeopenmixtureofexpertslanguage}
Muennighoff, N., Soldaini, L., Groeneveld, D., Lo, K., Morrison, J., Min, S., Shi, W., Walsh, P., Tafjord, O., Lambert, N., Gu, Y., Arora, S., Bhagia, A., Schwenk, D., Wadden, D., Wettig, A., Hui, B., Dettmers, T., Kiela, D., Farhadi, A., Smith, N.~A., Koh, P.~W., Singh, A., and Hajishirzi, H.
\newblock Olmoe: Open mixture-of-experts language models, 2024.
\newblock URL \url{https://arxiv.org/abs/2409.02060}.

\bibitem[Pearce \& Song(2024)Pearce and Song]{pearce2024reconcilingkaplanchinchillascaling}
Pearce, T. and Song, J.
\newblock Reconciling kaplan and chinchilla scaling laws, 2024.
\newblock URL \url{https://arxiv.org/abs/2406.12907}.

\bibitem[Penedo et~al.(2024)Penedo, Kydlíček, allal, Lozhkov, Mitchell, Raffel, Werra, and Wolf]{penedo2024finewebdatasetsdecantingweb}
Penedo, G., Kydlíček, H., allal, L.~B., Lozhkov, A., Mitchell, M., Raffel, C., Werra, L.~V., and Wolf, T.
\newblock The fineweb datasets: Decanting the web for the finest text data at scale, 2024.
\newblock URL \url{https://arxiv.org/abs/2406.17557}.

\bibitem[Porian et~al.(2025)Porian, Wortsman, Jitsev, Schmidt, and Carmon]{porian2025resolvingdiscrepanciescomputeoptimalscaling}
Porian, T., Wortsman, M., Jitsev, J., Schmidt, L., and Carmon, Y.
\newblock Resolving discrepancies in compute-optimal scaling of language models, 2025.
\newblock URL \url{https://arxiv.org/abs/2406.19146}.

\bibitem[Radford et~al.(2018)Radford, Narasimhan, Salimans, and Sutskever]{radford2018improving}
Radford, A., Narasimhan, K., Salimans, T., and Sutskever, I.
\newblock Improving language understanding by generative pre-training.
\newblock 2018.

\bibitem[Rae et~al.(2022)Rae, Borgeaud, Cai, Millican, Hoffmann, Song, Aslanides, Henderson, Ring, Young, Rutherford, Hennigan, Menick, Cassirer, Powell, van~den Driessche, Hendricks, Rauh, Huang, Glaese, Welbl, Dathathri, Huang, Uesato, Mellor, Higgins, Creswell, McAleese, Wu, Elsen, Jayakumar, Buchatskaya, Budden, Sutherland, Simonyan, Paganini, Sifre, Martens, Li, Kuncoro, Nematzadeh, Gribovskaya, Donato, Lazaridou, Mensch, Lespiau, Tsimpoukelli, Grigorev, Fritz, Sottiaux, Pajarskas, Pohlen, Gong, Toyama, de~Masson~d'Autume, Li, Terzi, Mikulik, Babuschkin, Clark, de~Las~Casas, Guy, Jones, Bradbury, Johnson, Hechtman, Weidinger, Gabriel, Isaac, Lockhart, Osindero, Rimell, Dyer, Vinyals, Ayoub, Stanway, Bennett, Hassabis, Kavukcuoglu, and Irving]{rae2022scaling}
Rae, J.~W., Borgeaud, S., Cai, T., Millican, K., Hoffmann, J., Song, F., Aslanides, J., Henderson, S., Ring, R., Young, S., Rutherford, E., Hennigan, T., Menick, J., Cassirer, A., Powell, R., van~den Driessche, G., Hendricks, L.~A., Rauh, M., Huang, P.-S., Glaese, A., Welbl, J., Dathathri, S., Huang, S., Uesato, J., Mellor, J., Higgins, I., Creswell, A., McAleese, N., Wu, A., Elsen, E., Jayakumar, S., Buchatskaya, E., Budden, D., Sutherland, E., Simonyan, K., Paganini, M., Sifre, L., Martens, L., Li, X.~L., Kuncoro, A., Nematzadeh, A., Gribovskaya, E., Donato, D., Lazaridou, A., Mensch, A., Lespiau, J.-B., Tsimpoukelli, M., Grigorev, N., Fritz, D., Sottiaux, T., Pajarskas, M., Pohlen, T., Gong, Z., Toyama, D., de~Masson~d'Autume, C., Li, Y., Terzi, T., Mikulik, V., Babuschkin, I., Clark, A., de~Las~Casas, D., Guy, A., Jones, C., Bradbury, J., Johnson, M., Hechtman, B., Weidinger, L., Gabriel, I., Isaac, W., Lockhart, E., Osindero, S., Rimell, L., Dyer, C., Vinyals, O., Ayoub, K., Stanway, J., Bennett, L.,
  Hassabis, D., Kavukcuoglu, K., and Irving, G.
\newblock Scaling language models: Methods, analysis and insights from training gopher, 2022.

\bibitem[Sardana et~al.(2024)Sardana, Portes, Doubov, and Frankle]{sardana2024chinchillaoptimalaccountinginferencelanguage}
Sardana, N., Portes, J., Doubov, S., and Frankle, J.
\newblock Beyond chinchilla-optimal: Accounting for inference in language model scaling laws, 2024.
\newblock URL \url{https://arxiv.org/abs/2401.00448}.

\bibitem[Shazeer(2020)]{shazeer2020gluvariantsimprovetransformer}
Shazeer, N.
\newblock Glu variants improve transformer, 2020.
\newblock URL \url{https://arxiv.org/abs/2002.05202}.

\bibitem[Shazeer et~al.(2017)Shazeer, Mirhoseini, Maziarz, Davis, Le, Hinton, and Dean]{shazeer2017outrageously}
Shazeer, N., Mirhoseini, A., Maziarz, K., Davis, A., Le, Q., Hinton, G., and Dean, J.
\newblock Outrageously large neural networks: The sparsely-gated mixture-of-experts layer, 2017.

\bibitem[Shazeer et~al.(2018)Shazeer, Cheng, Parmar, Tran, Vaswani, Koanantakool, Hawkins, Lee, Hong, Young, Sepassi, and Hechtman]{shazeer2018meshtensorflow}
Shazeer, N., Cheng, Y., Parmar, N., Tran, D., Vaswani, A., Koanantakool, P., Hawkins, P., Lee, H., Hong, M., Young, C., Sepassi, R., and Hechtman, B.
\newblock Mesh-tensorflow: Deep learning for supercomputers, 2018.

\bibitem[Su et~al.(2023)Su, Lu, Pan, Murtadha, Wen, and Liu]{su2023roformerenhancedtransformerrotary}
Su, J., Lu, Y., Pan, S., Murtadha, A., Wen, B., and Liu, Y.
\newblock Roformer: Enhanced transformer with rotary position embedding, 2023.
\newblock URL \url{https://arxiv.org/abs/2104.09864}.

\bibitem[Team(2024{\natexlab{a}})]{qwen25}
Team, Q.
\newblock Qwen2.5 technical report.
\newblock \emph{arXiv preprint arXiv:2412.15115}, 2024{\natexlab{a}}.

\bibitem[Team(2024{\natexlab{b}})]{qwen_moe}
Team, Q.
\newblock Qwen1.5-moe: Matching 7b model performance with 1/3 activated parameters", February 2024{\natexlab{b}}.
\newblock URL \url{https://qwenlm.github.io/blog/qwen-moe/}.

\bibitem[Yun et~al.(2024)Yun, Zhuang, Fu, Xing, and Zhang]{yun2024inferenceoptimalmixtureofexpertlargelanguage}
Yun, L., Zhuang, Y., Fu, Y., Xing, E.~P., and Zhang, H.
\newblock Toward inference-optimal mixture-of-expert large language models, 2024.
\newblock URL \url{https://arxiv.org/abs/2404.02852}.

\bibitem[Zadouri et~al.(2023)Zadouri, {\"U}st{\"u}n, Ahmadian, Ermi{\c{s}}, Locatelli, and Hooker]{zadouri2023pushing}
Zadouri, T., {\"U}st{\"u}n, A., Ahmadian, A., Ermi{\c{s}}, B., Locatelli, A., and Hooker, S.
\newblock Pushing mixture of experts to the limit: Extremely parameter efficient moe for instruction tuning.
\newblock \emph{arXiv preprint arXiv:2309.05444}, 2023.

\bibitem[Zoph et~al.(2022)Zoph, Bello, Kumar, Du, Huang, Dean, Shazeer, and Fedus]{zoph2022st}
Zoph, B., Bello, I., Kumar, S., Du, N., Huang, Y., Dean, J., Shazeer, N., and Fedus, W.
\newblock St-moe: Designing stable and transferable sparse expert models.
\newblock \emph{arXiv preprint arXiv:2202.08906}, 2022.

\end{thebibliography}
\bibliographystyle{icml2025}

\newpage
\appendix
\onecolumn

\section{Technical Details} \label{app:technical}

\subsection{Counting Parameters}
There are several ways to measure the size of a model. The two most important distinctions are whether total or active parameters are counted, and whether the parameters in the embedding and unembedding layers are included. Various papers assume different notations; notably, \citet{kaplan2020scaling} use nonembedding parameters, while \citet{hoffmann2022training} opt for the parameter count including embedding and unembedding.
Throughout our work, we try to make it clear which way of counting we are using in each particular instance. When no additional information is given, $\Nact$ and $\Ntotal$ denote respectively active and total parameters, including the embedding and unembedding.

If we let $\dm$ be the hidden dimension of the model and $\dv$ be the vocabulary size ($50,257$ in our case), then the following relations hold:
\begin{align}
    \Ntotal &= 2\dm\dv + (4 + 9E)N_{\text{blocks}}\dm^2 \\
    \Nact &= 2\dm\dv + 13N_{\text{blocks}}\dm^2
\end{align}

\subsection{Counting FLOPs}
Based on \citet{sardana2024chinchillaoptimalaccountinginferencelanguage}, we assume the cost of training to be $F_{\text{training}} = 6\Nact  D_{\text{training}}$, and the cost of inference to be $F_{\text{inference}} = 2\Nact D_{\text{inference}}$. Due to the relatively small number ($\leq 32$) of experts used with implicit expert granularity of $1.0$ \citep{ludziejewskiscaling}, we can consider the memory and FLOPs cost of routing to be negligible, following \citet{clark2022unified}.
\subsection{Model Configs}
The vast majority of our experiments use a simple rule for scaling the configuration, i.e., $N_{\text{blocks}} = N_{\text{heads}} = \text{d}_\text{model} / 64$ and assume these relations hold in all calculations. We base this rule on findings by \citet{kaplan2020scaling}.

\section{Fit Details} \label{app:fit_details}

\begin{table*}[h]
  \centering
\begin{tabular}{cccccccccccc}
\toprule
 $a$       & ${\alpha}$ & $\delta$   & $\gamma$  & $b$       & ${\beta}$ & $\omega$  & $\zeta$  & $E_{\text{start}}$ & $E_{\text{max}}$ & $c$        \\
\midrule
$35.91$     & $-0.1889$    & $-0.2285$     & $0.0098$   & $35.98$    & $-0.1775$ & $0.5529$    & $-0.0259$  & $2.0732$      & $290.4521$  & $1.3637$ \\

\bottomrule
\end{tabular}
\vspace{0.1cm}
  \label{tab:scaling_law_params}
  \caption{Fitted coefficients of our joined formula.}
\end{table*}

\begin{table*}[h]
  \centering
\begin{tabular}{r|lllll}
\toprule
$E$ & $m$ & $\mu$ & $n$ & $\nu$ & $c$ \\
\midrule
1  & \round{30.363993648167263} & \round{-0.18173956204827252} & \round{53.98384414155987 } &  \round{-0.19650191228646036}  & \round{1.3637} \\
2  & \round{27.79821195271775}  & \round{-0.17795397781692185} & \round{66.84011296802187 } &  \round{-0.20648914813917155}  & \round{1.3637} \\
4  & \round{24.846202931338134} & \round{-0.17314011944591995} & \round{87.70219698814259 } &  \round{-0.21918920603633535}  & \round{1.3637} \\
8  & \round{21.832989774298056} & \round{-0.1675966335114486 } & \round{119.91258457347162} &  \round{-0.23381418809729415}  & \round{1.3637} \\
16 & \round{19.015896366339852} & \round{-0.16167306968397718} & \round{167.50725788774776} &  \round{-0.24944190245208414}  & \round{1.3637} \\
32 & \round{16.54244596529427}  & \round{-0.1556980993088928 } & \round{234.67256254388218} &  \round{-0.2652052390210445 }  & \round{1.3637} \\
\bottomrule
\end{tabular}
\caption{The fitted coefficients of our joint formula, Equation~\ref{eq:scaling_law_final}, reduced to the Chinchilla scaling law, Equation~\ref{eq:scaling_law_chinchilla}, for a given number of experts, $E$. We observe that the dataset exponent, $\nu$, increases significantly. This is one of the reasons why compute-optimal parameter-to-token ratios change with $E$. }
\end{table*}

Following \citet{hoffmann2022training}, we use the LBFGS algorithm with a learning rate of $1e{-4}$ and a weight decay of $1e{-5}$ to fit the coefficients of Equation~\ref{eq:scaling_law_final}, optimizing the Huber loss with $\delta = 0.01$ over the set of our training runs described in the table in Appendix~\ref{sec:full_experiments}. Instead of removing outliers and underperforming models from the training set, we underweight them proportionally to the loss. Optimization hyperparameters were manually tuned to minimize error over the training dataset. The final fitted coefficients of Equation~\ref{eq:scaling_law_final} are within the boundaries of the grid of initializations given by: 
$\alpha \in \{0.05, 0.25, 0.5\}$,  
$\beta \in \{0.05, 0.25, 0.5\}$,  
$A \in \{30, 100, 300\}$,  
$B \in \{30, 100, 300\}$,  
$C \in \{0.5, 1, 2\}$,  
$\delta \in \{-0.5, 0, 0.5\}$,  
$\gamma \in \{-0.5, 0, 0.5\}$,  
$\omega \in \{-0.5, 0, 0.5\}$,  
$\zeta \in \{-0.5, 0, 0.5\}$. The selected coefficients were those with the lowest score, defined as the sum of RMSE on the training and a held-out extrapolation validation set. The formula in Equation~\ref{eq:scaling_law_final} was calculated in logarithm, without any exponentials, using only linear transformations and the logsumexp operation. It was optimized to predict the logarithm of $L$, and parameters $a$, $b$, and $c$ were optimized in logarithm. All these steps were taken to increase numerical stability and were essential for proper convergence.

\section{Compute- \& Memory-Matched Models}

\begin{figure*}[h]
    \begin{center}
        \includegraphics[width=\textwidth]{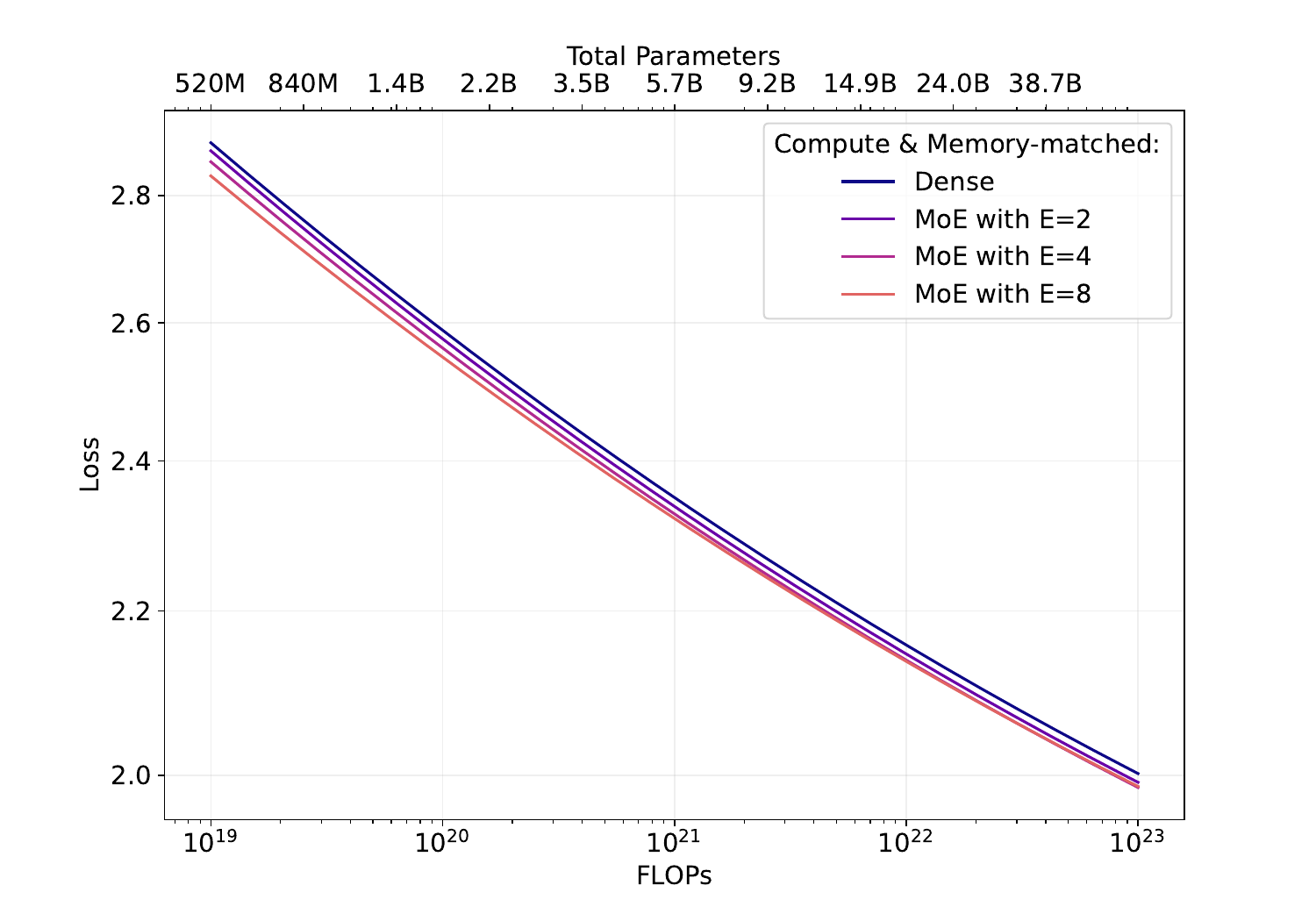}
    \end{center}
    \caption{Comparison between compute- and memory-matched models with different values of $E$. The corresponding total memory constraint for MoE models is derived from the compute-optimal model size for the dense model. 
    Due to the nature of this constraint, we do not consider higher values of $E$, as their token-to-parameter ratio significantly exceeds the threshold within which we believe our scaling law applies. For instance, an MoE model with $E=16$ that matches a $1$B dense model trained on $10$B tokens in FLOPs and memory would have $155$M activated parameters trained on $64$B tokens. This results in a token-to-parameter ratio of approximately $414$, surpassing the range covered by our dataset.}
    \label{plot:com_and_mem_matched}
\end{figure*}

\newpage
\section{Learning Rate Scaling Fit}\label{app:lr_scaling}

\begin{figure}[h]
    \begin{center}
        \includegraphics[width=\textwidth]{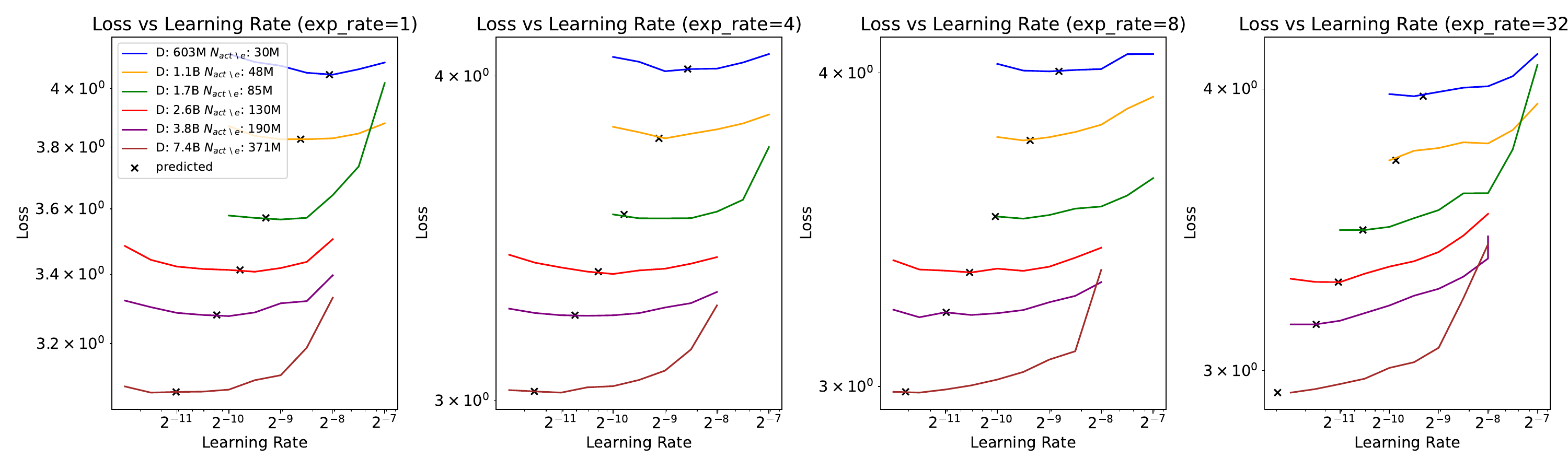}
    \end{center}
    \caption{Visualization of the fit ($E \in \{1,8\}$) of our LR scaling rule, interpolation ($E=4$) and extrapolation ($E=32$).}
    \label{plot:lr_scaling_with_e}
\end{figure}

\begin{figure}[h]
    \begin{center}
        \includegraphics[width=\textwidth]{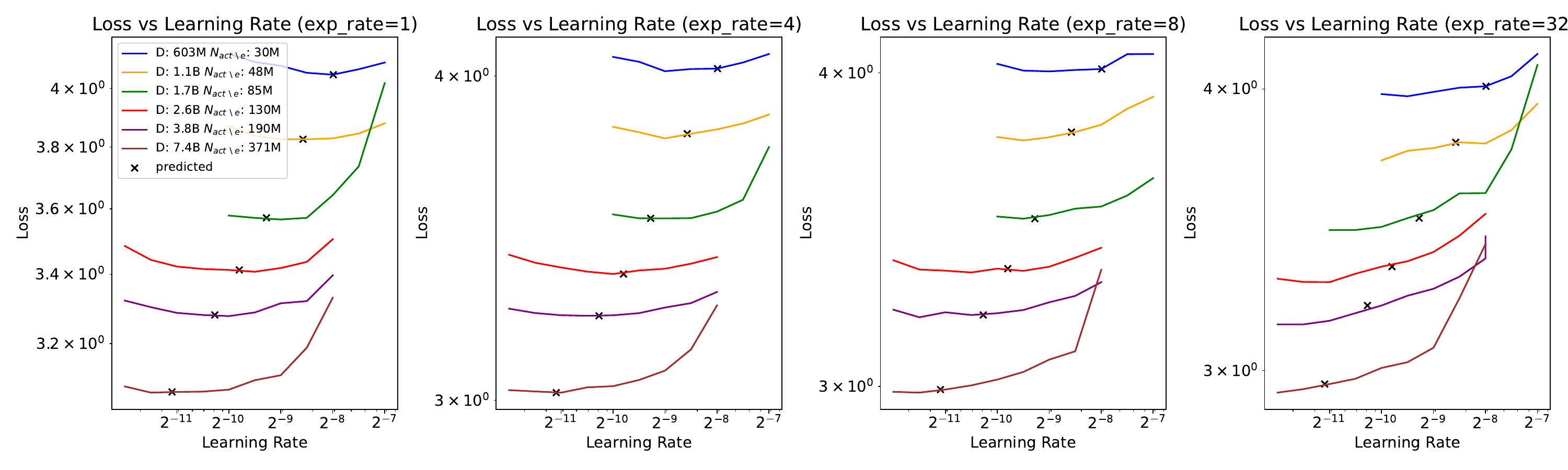}
    \end{center}
    \caption{Ablation for the LR scaling rule fit without considering the number of experts $E$. While performance on the training set ($E \in \{1,8\}$) looks acceptable, the extrapolation at $E=32$ is clearly suboptimal, validating the need for considering $E$.}
  \label{plot:lr_scaling_no_e}
\end{figure}

\section{Experiments Listing}
\label{sec:full_experiments}
{
\small
\begin{tabular}{lrrrlrl}
\toprule
$\Ntotal$ & $\text{N}_{\text{attn\_heads}}$ & $\text{N}_{\text{blocks}}$ & $\dm$ & $\Nact$ & $E$ & $D$ \\
\midrule
5.0B & 16 & 16 & 1024 & 321M & 32 & 16.0B, 8.0B, 4.0B, 2.0B, 1.0B, 500M \\
3.8B & 28 & 28 & 1792 & 1.3B & 4 & 11.1B, 5.6B, 2.8B, 2.0B \\
3.3B & 11 & 21 & 1408 & 683M & 8 & 16.0B, 8.0B, 4.0B, 2.0B, 1.0B, 500M \\
3.0B & 26 & 26 & 1664 & 1.1B & 4 & 80.0B, 64.0B, 48.0B, 32.0B, 16.0B, 8.0B, 4.0B, 2.0B, 1.0B, 500M \\
2.7B & 36 & 36 & 2304 & 2.7B & 1 & 9.2B, 5.5B, 2.8B, 2.0B, 1.4B, 980M \\
2.6B & 30 & 30 & 1920 & 1.6B & 2 & 5.4B, 2.7B \\
2.6B & 16 & 16 & 1024 & 321M & 16 & 16.0B, 8.0B, 4.0B, 2.0B, 1.0B, 500M \\
2.2B & 28 & 28 & 1792 & 1.3B & 2 & 18.6B, 11.1B, 5.6B, 4.0B, 2.8B, 2.0B \\
2.1B & 12 & 12 & 768 & 169M & 32 & 8.0B, 4.0B, 2.0B, 1.0B, 500M \\
2.1B & 10 & 16 & 1280 & 469M & 8 & 32.0B, 16.0B, 8.0B, 4.0B, 2.0B, 1.0B \\
1.9B & 22 & 22 & 1408 & 709M & 4 & 35.3B, 12.2B, 10.6B, 7.7B, 5.3B, 3.8B \\
1.8B & 11 & 21 & 1408 & 683M & 4 & 8.0B, 16.0B, 4.0B, 2.0B, 1.0B, 500M \\
1.8B & 26 & 26 & 1664 & 1.1B & 2 & 16.0B, 8.0B, 4.0B, 2.0B, 1.0B, 500M \\
1.6B & 30 & 30 & 1920 & 1.6B & 1 & 5.4B, 2.7B \\
1.4B & 16 & 16 & 1024 & 321M & 8 & 16.0B, 8.0B, 4.0B, 2.0B, 1.0B, 500M \\
1.3B & 28 & 28 & 1792 & 1.3B & 1 & 6.5B, 3.3B, 18.6B, 11.1B, 5.6B, 4.0B, 2.8B, 2.0B \\
1.3B & 10 & 10 & 640 & 118M & 32 & 4.0B, 2.0B, 1.0B, 500M \\
1.2B & 10 & 16 & 1280 & 469M & 4 & 32.0B, 16.0B, 8.0B, 4.0B, 2.0B, 1.0B, 500M \\
1.1B & 12 & 12 & 768 & 169M & 16 & 8.0B, 4.0B, 2.0B, 1.0B, 500M \\
1.1B & 26 & 26 & 1664 & 1.1B & 1 & 14.0B, 12.0B, 10.0B, 80.0B, 64.0B, 48.0B, 32.0B \\
1.1B & 26 & 26 & 1664 & 1.1B & 1 & 16.0B, 8.0B, 4.0B, 2.0B, 1.0B, 500M \\
1.1B & 22 & 22 & 1408 & 709M & 2 & 3.8B, 49.8B, 24.9B, 12.5B, 6.2B, 3.1B, 1.6B, 778M \\
1.1B & 22 & 22 & 1408 & 709M & 2 & 21.8B, 18.7B, 15.6B, 35.3B, 12.2B, 10.6B, 7.7B, 5.3B \\
1.1B & 18 & 18 & 1152 & 426M & 4 & 31.0B, 25.9B, 20.7B, 10.4B, 5.2B, 2.6B, 1.3B \\
1.1B & 11 & 21 & 1408 & 683M & 2 & 32.0B, 16.0B, 8.0B, 4.0B, 2.0B, 1.0B, 500M \\
890M & 24 & 24 & 1536 & 890M & 1 & 9.9B, 5.0B \\
850M & 20 & 20 & 1280 & 555M & 2 & 16.0B, 8.0B \\
774M & 16 & 16 & 1024 & 321M & 4 & 16.0B, 8.0B, 4.0B, 2.0B, 1.0B, 500M \\
709M & 22 & 22 & 1408 & 709M & 1 & 35.3B, 12.2B, 10.6B, 7.7B, 5.3B, 3.8B, 12.5B, 6.2B \\
705M & 10 & 16 & 1280 & 469M & 2 & 32.0B, 16.0B, 8.0B, 4.0B, 2.0B, 1.0B, 500M \\
683M & 11 & 21 & 1408 & 683M & 1 & 32.0B, 16.0B, 8.0B, 4.0B, 2.0B, 1.0B, 500M \\
671M & 10 & 10 & 640 & 118M & 16 & 4.0B, 2.0B, 1.0B, 500M \\
664M & 8 & 8 & 512 & 79M & 32 & 2.0B, 1.0B, 500M \\
615M & 12 & 12 & 768 & 169M & 8 & 8.0B, 4.0B, 2.0B, 1.0B, 500M \\
555M & 20 & 20 & 1280 & 555M & 1 & 16.0B, 8.0B \\
472M & 16 & 16 & 1024 & 321M & 2 & 16.0B, 8.0B, 4.0B, 2.0B, 1.0B, 500M \\
469M & 10 & 16 & 1280 & 469M & 1 & 32.0B, 16.0B, 8.0B, 4.0B, 2.0B, 1.0B, 500M \\
376M & 10 & 10 & 640 & 118M & 8 & 4.0B, 2.0B, 1.0B, 500M \\
362M & 8 & 8 & 512 & 79M & 16 & 2.0B, 1.0B, 500M \\
360M & 12 & 12 & 768 & 169M & 4 & 8.0B, 4.0B, 2.0B, 1.0B, 500M \\
321M & 16 & 16 & 1024 & 321M & 1 & 16.0B, 8.0B, 4.0B, 2.0B, 1.0B, 500M \\
289M & 11 & 11 & 704 & 142M & 4 & 4.5B, 2.3B, 1.1B \\
285M & 9 & 9 & 576 & 97M & 8 & 3.3B, 1.7B \\
282M & 13 & 13 & 832 & 201M & 2 & 6.4B, 3.2B, 1.6B, 800M \\
233M & 12 & 12 & 768 & 169M & 2 & 8.0B, 4.0B, 2.0B, 1.0B, 500M \\
228M & 10 & 10 & 640 & 118M & 4 & 4.0B, 2.0B, 1.0B, 500M \\
211M & 8 & 8 & 512 & 79M & 8 & 2.0B, 1.0B, 500M \\
169M & 12 & 12 & 768 & 169M & 1 & 8.0B, 4.0B, 2.0B, 1.0B, 500M \\
154M & 10 & 10 & 640 & 118M & 2 & 4.0B, 2.0B, 1.0B, 500M \\
135M & 8 & 8 & 512 & 79M & 4 & 2.0B, 1.0B, 500M \\
118M & 10 & 10 & 640 & 118M & 1 & 4.0B, 2.0B, 1.0B, 500M \\
98M & 8 & 8 & 512 & 79M & 2 & 2.0B, 1.0B, 500M \\
79M & 8 & 8 & 512 & 79M & 1 & 2.0B, 1.0B, 500M \\
\bottomrule
\end{tabular}
}

\end{document}